\documentclass{article}

\PassOptionsToPackage{numbers, compress}{natbib}

\usepackage[main, final]{neurips_2026}

\usepackage[utf8]{inputenc} %
\usepackage[T1]{fontenc}    %
\usepackage{hyperref}       %
\usepackage{url}            %
\usepackage{booktabs}       %
\usepackage{amsfonts}       %
\usepackage{nicefrac}       %
\usepackage{microtype}      %
\usepackage{xcolor}         %

\usepackage{amsmath,amssymb,amsthm,mathtools}
\usepackage{hyperref}
\usepackage{comment}
\usepackage{nicefrac}
\usepackage{tikz}
\usetikzlibrary{arrows.meta}
\usepackage{algorithm}
\usepackage{algpseudocode}

\newcommand{\R}{\mathbb{R}}
\newcommand{\E}{\mathbb{E}}
\newcommand{\PP}{\mathbb{P}}

\newcommand{\sign}{\operatorname{sign}}
\newcommand{\Var}{\operatorname{Var}}
\newcommand{\Cov}{\operatorname{Cov}}

\newcommand{\x}{\mathbf{x}}
\newcommand{\err}{\operatorname{err}}

\newcommand{\ph}{\phi}

\newcommand{\cN}{\mathcal{N}}

\newif\ifshowcomments
\showcommentsfalse

\usepackage{xcolor}

\theoremstyle{definition}
\newtheorem{definition}{Definition}[section]
\newtheorem{remark}[definition]{Remark}

\theoremstyle{plain}
\newtheorem{theorem}[definition]{Theorem}
\newtheorem{lemma}[definition]{Lemma}
\newtheorem{proposition}[definition]{Proposition}
\newtheorem{corollary}[definition]{Corollary}

\title{Rare-Tail Statistics For Learning Biased Gaussian Halfspaces with Label Noise}

\author{%
  Arya Mazumdar \\ Halıcıoğlu School of \\ Data Science and Computing \\ UC San Diego \\ \texttt{arya@ucsd.edu}
  \And
  Prateeti Mukherjee \\ Halıcıoğlu School of \\ Data Science and Computing \\ UC San Diego \\ \texttt{pmukherjee@ucsd.edu}
  \And
  Harsh Vardhan \\ Computer Science \\ and Engineering \\ UC San Diego \\ \texttt{hharshvardhan@ucsd.edu}
}

\renewcommand{\epsilon}{\varepsilon}
\begin{document}

\maketitle

\begin{abstract}
We revisit the classical problem of learning general Gaussian halfspaces in the presence of label noise, with particular focus on the large-bias regime, where information-computation gaps are conjectured in several standard label-noise models. Concretely, we present and analyze the rare-tail mean estimator, an extremely simple one-shot statistic that averages only the observed minority class. In the large-bias regime, classical Averaging (Servedio, 1999) and related global first-moment statistics are provably suboptimal, while the rare-tail mean estimator recovers the target halfspace direction at the information-theoretic sample rate (modulo log factors) under a broad class of label noise models, including probit noise and margin-dependent noise with very slow decay away from the boundary. Notably, our guarantees allow labels arbitrarily close to the decision boundary to be nearly random, while still giving strong directional recovery from a small number of samples. The estimator also cleanly interpolates between the noiseless and noisy regimes: without noise, it attains the same sample rate as classical separability-based methods for realizable halfspaces, and under low noise, or noise that decays away from the boundary, it keeps this rate even after linear separability has been destroyed. As an application, we consider a two-stage curriculum in which source samples, whose noise may be arbitrarily close to $1/2$ near the boundary and decay very slowly away from it, precede target samples, whose noise flips every label at the same rate, i.e., random classification noise (RCN). Since both stages share the same Bayes optimal halfspace, our estimator supplies the directional warm start needed by existing refinement routines, and together they learn the RCN target at the information-theoretic sample rate in the large-bias regime, up to log factors.

\end{abstract}

\section{Introduction}

A recurring theme in learning theory is that provable guarantees can be highly sensitive to the structure of label noise. The classical random classification noise model (RCN), in which every label is independently flipped at a fixed rate, is straightforward to analyze, and has played a foundational role in the theory of noisy halfspace learning~\citep{angluin1988learning}. However, even mild departures from this model can change the statistical and computational behavior of classification problems. For halfspaces, allowing the corruption rate to vary with the instance, or with the margin to the decision boundary, has led to substantially different algorithms and strong statistical-query lower bounds~\citep{diakonikolas2022nearoptimalmassart,nasser2022optimal,diakonikolas2021tsybakov}. In this work we revisit a particularly sharp instance of this phenomenon: learning an affine Gaussian halfspace under label noise in the large-bias regime.

Concretely, we observe i.i.d. labeled examples $(X,Y)$ with $X\sim\cN(0,I_d)$, $Y\in\{\pm1\}$, and an unknown target separator $f^\star(x)=\sign(m(x))$, where $m(x)=\langle w^\star,x\rangle+t^\star$ is the signed margin, and $w^\star \in S^{d-1}, t^\star \in \R$. Our goal is to output a halfspace whose excess risk over $f^\star$ is at most a target accuracy $\varepsilon$. 

For ease of exposition, assume the threshold $t^\star >0$ (the same ideas apply to $t^\star < 0$ by symmetry). Then, the halfspace has bias $p = \min\{\PP(f^\star(X)=-1), \PP(f^\star(X)=+1)\} = \Phi(-t^\star)$, where $\Phi(\cdot)$ is the usual Gaussian CDF. Conditioned on $X=x$, the observed label satisfies $Y=f^\star(x)$ with probability $1-g(|m(x)|)$ and $Y=-f^\star(x)$ with probability $g(|m(x)|)$, where $g:[0,\infty)\to[0,\nicefrac{1}{2}]$ is nonincreasing. This margin-dependent noise profile, going back to Bylander's \cite{bylander1998learning} monotonic-noise model for linear threshold functions, interpolates between the flat $\eta$-RCN model, where $g(\cdot)=\eta$ and every label is flipped at the same rate, and boundary-local noise, where corruption is concentrated near the decision boundary and decays away from it. The latter describes a realistic annotation regime, in which labels far from the boundary are more reliable, while labels near the boundary may be close to random. Crowdsourced labels, for example from Amazon Mechanical Turk (AMT), behave this way. Annotators tend to agree on prototypical instances, which we can think of as points far from the decision boundary, and their labels become nearly random on ambiguous instances close to the separator~\cite{whitehill2009whose,beigman2009}. In contrast, $\eta$-RCN describes errors that do not depend on the instance at all, such as labels from annotators who answer without looking at the samples~\cite{raykar}.

The large-bias regime is where the distinction between flat and boundary-local noise becomes most pronounced. When $t^\star>0$ is large, the true negative class has mass $p=\Phi(-t^\star)\ll 1$, and most samples lie deep in the positive region. Conditioning on the positive side removes only a set of mass $p$, so a positive sample looks almost like a draw from $\cN(0,I_d)$ and says little about the orientation of the separator. The negative samples, on the other hand, all come from the thin Gaussian tail cut off by the separator, so where they lie is tied directly to the orientation of the halfspace. In the absence of noise, conditioning on this tail is extremely favorable: if $X=Z w^\star+X_\perp$ with $Z = \langle w^\star, X\rangle \sim\cN(0,1)$ and $X_\perp\sim\cN(0, I_d - w^\star w^{\star T})$, then the event $Y=-1$ is simply $\{Z<-t^\star\}$. Conditioning on this event then shifts the $Z$ coordinate, but leaves the orthogonal directions unchanged in distribution. Consequently, the signal $\E[X\mid Y=-1]$ is collinear with $-w^\star$ and has large norm $\Theta(t^\star)$ relative to the unit variance noise in every direction.  

\begin{figure}[t]
\centering
\def\Ax{-0.52}
\def\Ay{-0.86}
\begin{minipage}[t]{0.48\textwidth}
\centering
\begin{tikzpicture}[x=1.4cm,y=1.4cm]

    \coordinate (O) at (0,0);
    \coordinate (A) at (\Ax,\Ay);

    \draw[gray!55, densely dashed] (-1.03,-1.71) -- (1.03,1.71);     %
    \draw[gray!55, densely dashed] (-1.80,1.08) -- (1.80,-1.08);     %

    \draw[->, thin] (-2.35,0) -- (2.05,0) node[right] {$x_1$};
    \draw[->, thin] (0,-2.35) -- (0,1.75) node[above] {$x_2$};

    \draw[gray!55] (0,0) circle (0.65);
    \draw[gray!55] (0,0) circle (1.20);
    \draw[gray!55] (0,0) circle (1.75);
    \node[gray!70] at (-1.25,1.35) {\small $\mathcal N(0,I_2)$};

    \draw[very thick] (-2.20,0.15) -- (1.55,-2.10);

    \fill[black] (O) circle (1.4pt);
    \fill[black] (A) circle (1.4pt);
    \draw[<->, black!75, line width=0.9pt]   (O) -- (A) node[midway, left] {\scriptsize $t^\star$};
    \node[black, below left] at (A) {\scriptsize $-t^\star\,w^\star$};

    \draw[red!80!black, very thick, dashed] (-1.79,1.05) -- (0.76,-2.77);

    \foreach \x/\y in {
        -1.12/0.08,
        -0.95/0.72,
        -0.61/-0.16,
        -0.42/0.47,
        -0.18/0.03,
        -0.03/0.92,
         0.17/-0.31,
         0.34/0.24,
         0.52/0.77,
         0.70/-0.04,
         0.88/0.43,
         1.03/0.96,
         1.18/0.16,
         1.42/0.68,
         1.56/-0.21,
         0.24/1.24,
        -0.51/1.05
    }{
        \fill[blue!70!black] (\x,\y) circle (1.6pt);
    }

    \foreach \x/\y in {
        -1.86/-1.01,
        -1.52/-1.28,
        -1.35/-0.74,
        -1.08/-1.00
    }{
        \fill[orange!85!black] (\x,\y) circle (1.8pt);
    }

    \draw[->, line width=1.1pt, black] (0,0) -- (0.90,1.50);
    \node[black, right] at (0.90,1.50) {\small $w^\star$};

    \draw[->, line width=0.9pt, red!80!black] (0,0) -- (0.85,0.57);
    \node[red!80!black, right] at (0.85,0.57) {\small $\widehat\mu_{\mathrm{avg}}$};

    \node[blue!70!black] at (1.15,0.15) {\scriptsize majority region};
    \node[orange!85!black] at (-1.65,-1.48) {\scriptsize minority tail};

    \node[font=\bfseries] at (0,-2.60) {(a) Average estimator};
\end{tikzpicture}
\end{minipage}
\hfill
\begin{minipage}[t]{0.48\textwidth}
\centering
\begin{tikzpicture}[x=1.4cm,y=1.4cm]

    \coordinate (O) at (0,0);
    \coordinate (A) at (\Ax,\Ay);

    \draw[gray!55, densely dashed] (-1.03,-1.71) -- (1.03,1.71);     %
    \draw[gray!55, densely dashed] (-1.80,1.08) -- (1.80,-1.08);     %

    \draw[->, thin] (-2.35,0) -- (2.05,0) node[right] {$x_1$};
    \draw[->, thin] (0,-2.35) -- (0,1.75) node[above] {$x_2$};

    \draw[gray!55] (0,0) circle (0.65);
    \draw[gray!55] (0,0) circle (1.20);
    \draw[gray!55] (0,0) circle (1.75);
    \node[gray!70] at (-1.25,1.35) {\small $\mathcal N(0,I_2)$};

    \draw[very thick] (-2.20,0.15) -- (1.55,-2.10);

    \fill[black] (O) circle (1.4pt);
    \fill[black] (A) circle (1.4pt);
    \draw[<->, black!75, line width=0.9pt]   (O) -- (A) node[midway, left] {\scriptsize $t^\star$};
    \node[black, below left] at (A) {\scriptsize $-t^\star\,w^\star$};

    \draw[teal!70!black, very thick, dashed] (-2.32,0.57) -- (1.29,-2.29);

    \foreach \x/\y in {
        -1.12/0.08,
        -0.95/0.72,
        -0.61/-0.16,
        -0.42/0.47,
        -0.18/0.03,
        -0.03/0.92,
         0.17/-0.31,
         0.34/0.24,
         0.52/0.77,
         0.70/-0.04,
         0.88/0.43,
         1.03/0.96,
         1.18/0.16,
         1.42/0.68,
         1.56/-0.21,
         0.24/1.24,
        -0.51/1.05
    }{
        \fill[blue!70!black] (\x,\y) circle (1.6pt);
    }

    \foreach \x/\y in {
        -1.86/-1.01,
        -1.52/-1.28,
        -1.35/-0.74,
        -1.08/-1.00
    }{
        \fill[orange!85!black] (\x,\y) circle (1.8pt);
    }

    \draw[->, line width=1.1pt, black] (0,0) -- (0.90,1.50);
    \node[black, right] at (0.90,1.50) {\small $w^\star$};

    \draw[->, line width=1.1pt, teal!70!black] (0,0) -- (1.05,1.33);
    \node[teal!70!black, right] at (1.05,1.33) {\small $-\widehat{\mu}_-$};

    \node[blue!70!black] at (1.15,0.15) {\scriptsize majority region};
    \node[orange!85!black] at (-1.65,-1.48) {\scriptsize minority tail};

    \node[font=\bfseries] at (0,-2.60) {(b) Minority mean estimator};
\end{tikzpicture}
\end{minipage}
\caption{A 2D cartoon of the noiseless biased regime. The solid black line is the true boundary and the point $-t^\star w^\star$ is its closest point to the origin. The dashed gray axes depict the decomposition $X=Z\,w^\star+X_\perp$, and the colored dashed separators are drawn through the same point $-t^\star w^\star$ to isolate directional error. Panel (a) uses the empirical Average moment, while panel (b) uses the empirical minority mean. For the same sample budget $n=\tilde O(\nicefrac{d}{p})$, the minority mean typically achieves $\tilde O(\nicefrac{1}{t^\star})$-scale angular distance to $w^\star$, while the Average estimator typically only reaches $\tilde O(\nicefrac{1}{t^\star\sqrt{p}})$-scale, which explains the larger angle in (a) compared to (b). The signal magnitude of the Average moment is smaller by a $2p$ factor, hence the shorter arrow in (a), while its orthogonal noise is smaller only by a $\sqrt{p}$ factor, since it also sums over the majority samples.}
\label{fig:directional-overview}
\end{figure}
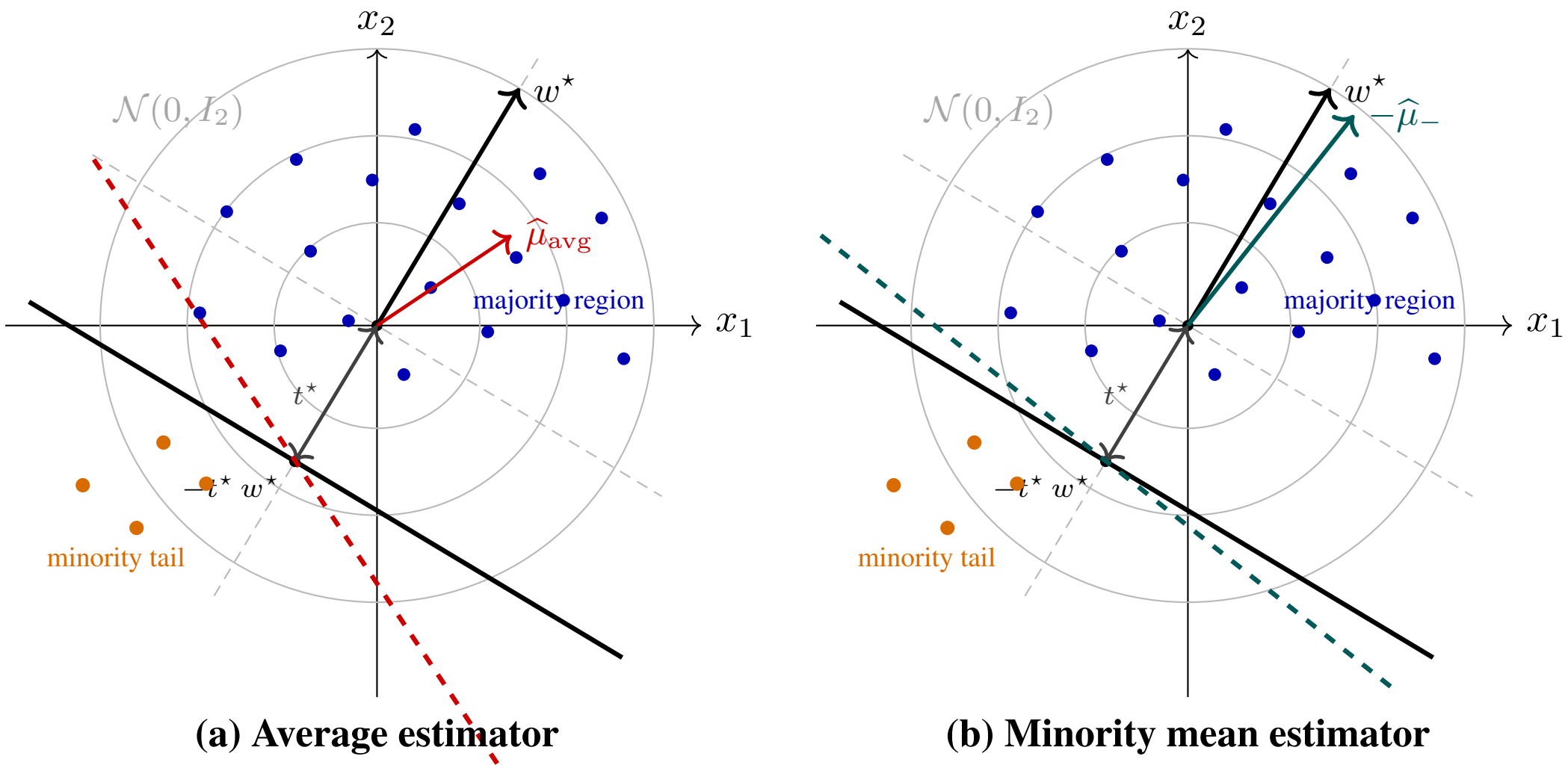

This suggests an extremely simple estimator: average the examples with the empirically rarer label and orient the resulting vector according to that label. We call this the rare-tail direction estimator (Algorithm~\ref{alg:rare-tail-direction}). This is essentially a variant of Servedio's classical Averaging algorithm \cite{Servedio1999OnPL}, with the labels centered at the majority label, so that the majority samples, which carry noise but no additional signal, drop out of the average. The difference is small for homogeneous halfspaces, but striking for affine ones: in the noiseless biased benchmark discussed above, this rare-tail statistic reaches $O(\nicefrac{1}{t^\star})$ angular precision at the information-theoretically optimal $\widetilde O(\nicefrac{d}{p})$ sample rate, whereas the global Average statistic requires $\widetilde O(\nicefrac{d}{p^2})$ samples to reach the same precision (Figure~\ref{fig:directional-overview}).%

It remains to understand whether this geometric intuition survives when the labels are unreliable. We find that the answer depends on the shape of the noise profile, as opposed to just the amount of noise there is. In particular, constant-rate flat noise profiles destroy the rare-tail advantage. Under $\eta$-RCN, points far inside the majority region are flipped into the observed negative class at the same rate as points near the boundary. In the large-bias regime, this creates a constant-rate stream of false negatives from the Gaussian bulk that dilute the signal, since the observed negative class is no longer a rare tail event, but a mixture of the true tail and mislabeled majority points. Boundary-local noise can produce just as many false negatives, even far more than the true tail, but these are drawn preferentially from near the separator, which, in the large-bias regime, lies far from the origin, in the direction $-w^\star$. Therefore, the observed negative class is no longer the true tail, yet the false negatives themselves pull its mean along $-w^\star$, thereby turning label noise into signal. The obstruction is therefore not inherent to label noise itself, but to the flatness of the noise profile.

\emph{The central question of the paper is whether this one-shot statistic retains its geometric advantage when the noise is not flat, but is otherwise as severe as possible.} %

We answer this question in the affirmative for a broad class of noise profiles that satisfy a mild regularity condition. This condition, which we call regular variation, rules out RCN-like flat plateaus, but includes noise models that may force random labels close to the decision boundary, but have very slow decay with distance to it. In particular, it includes canonical boundary-local models such as probit noise, which permits the flip probability near the boundary to be arbitrarily close to $\nicefrac{1}{2}$. Our main theorem shows that, under this condition, the rare-tail direction estimator recovers $w^\star$ to $O(\nicefrac{1}{t^\star})$ angular accuracy using the same leading sample scale %
as the noiseless regime. Moreover, this is exactly the angular scale needed in the large-bias regime. Indeed, when $p=\Theta(\epsilon)$, a direction within $O(\nicefrac{1}{t^\star})$ and a correctly chosen threshold already yield $\epsilon$ excess risk. A one-dimensional threshold search along the recovered direction (Algorithm~\ref{alg:threshold-erm}) then gives a complete learner that runs in nearly linear time.

Isolating the shape of the noise profile as the obstruction also sheds light on a known computational barrier for learning biased Gaussian halfspaces under RCN. \citet{diakonikolas2023nearoptimalboundslearninggaussian} give efficient algorithms for $\eta$-RCN, but in the critical regime $p=\Theta(\varepsilon)$, their sample complexity has a leading $\widetilde O(\nicefrac{d}{\varepsilon^2})$ term, while the information-theoretic rate is $\widetilde \Theta(\nicefrac{d}{\varepsilon})$. They also give a statistical-query lower bound, suggesting that this quadratic dependence is unavoidable for efficient learners that only see RCN samples. Our analysis suggests where this barrier comes from, namely, the search for a warm start in the presence of constant-rate flat noise. As discussed above, flat noise buries much of the signal, while the same halfspace is easy to warm-start from boundary-local noisy labels. This is also the only part where flatness meaningfully affects learning: given a direction within $O(\nicefrac{1}{t^\star})$ of $w^\star$, the threshold is a one-dimensional learning problem, and the local refinement procedure of \citet{diakonikolas2023nearoptimalboundslearninggaussian} from this warm-start is sample-efficient, even under RCN.

This suggests a way around the barrier: obtain the warm start from labels whose noise is not flat, and use the RCN samples only for the remaining steps. Strategies of this kind, in which a learner sees data in stages, starting from samples where the signal is easier to extract, have long been used in empirical machine learning under the name of curriculum learning \citep{bengio2009curriculum}, including for training on noisy labels, where the curriculum favors examples whose labels are more likely to be correct \citep{Guo_2018_ECCV, mentor_net}. Our two-stage approach serves as a mathematical model of curriculum learning for noisy halfspaces, where both stages share the Gaussian marginal and Bayes-optimal halfspace, but differ only in their label noise. The boundary-local source stage is not a uniformly easier version of the target, since near the boundary its labels may be even noisier than the target's. Instead, it resembles careful annotators, whose mistakes concentrate on ambiguous instances, as in the crowdsourcing example above. In contrast, the $\eta$-RCN target stage resembles a pool in which each label, with probability $2\eta$, comes from an annotator who answers at random, without looking at the instance. Our curriculum learner runs the rare-tail estimator on the source stage to obtain a warm start, and then uses target RCN samples to estimate the threshold and refine the direction locally, following \citet{diakonikolas2023nearoptimalboundslearninggaussian}. Since the learner also sees the source stage, the statistical-query lower bound does not apply, even though the final guarantee is measured on the RCN target.

\subsection{Our Results}
\label{subsec:our-results}

\begin{algorithm}[t]
\caption{\textsc{Rare-Tail Direction Estimator}}
\label{alg:rare-tail-direction}
\begin{algorithmic}[1]
\Require Samples $S=\{(X_i,Y_i)\}_{i=1}^N$ with $X_i\in\mathbb R^d$ and $Y_i\in\{\pm1\}$
\Ensure A unit vector $\widehat w\in\mathbb S^{d-1}$ and empirical rare label $\widehat y_{\rm rt}$
\State $N_+ \gets \sum_{i=1}^N \mathbf 1\{Y_i=+1\}$, \quad
       $N_- \gets \sum_{i=1}^N \mathbf 1\{Y_i=-1\}$
\State $\widehat y_{\rm rt}\gets -1$ if $N_-\le N_+$, and $\widehat y_{\rm rt}\gets +1$ otherwise
\State $I_{\rm rt}\gets \{i\in[N]:Y_i=\widehat y_{\rm rt}\}$
\If{$I_{\rm rt}=\emptyset$}
    \State \Return any fixed unit vector and $\widehat y_{\rm rt}$
\EndIf
\State $\widehat\mu_{\rm rt}\gets |I_{\rm rt}|^{-1}\sum_{i\in I_{\rm rt}}X_i$
\If{$\widehat\mu_{\rm rt}=0$}
    \State \Return any fixed unit vector and $\widehat y_{\rm rt}$
\EndIf
\State $\widehat w\gets \widehat y_{\rm rt}\widehat\mu_{\rm rt}/\|\widehat\mu_{\rm rt}\|_2$
\State \Return $(\widehat w,\widehat y_{\rm rt})$
\end{algorithmic}
\end{algorithm}

We summarize the main guarantees. Throughout, $p=\Phi(-t^\star)$ denotes the bias of the halfspace under the convention $t^\star>0$, and our primary regime is the critical large-bias regime $p=\Theta(\epsilon)$.

Our first result shows that the rare-tail direction estimator obtains the same leading sample scale as in the noiseless benchmark for a broad class of noise profiles that satisfy a mild regularity condition, which we call regular-variation (Definition~\ref{def:regularly-varying-margin-noise}). 
\begin{theorem}[Informal: rare-tail direction recovery]
\label{thm:informal-rv-direction}
Under regularly-varying margin-dependent noise, Algorithm~\ref{alg:rare-tail-direction} returns a direction $\widehat w$ with $\angle(\widehat w,w^\star)\le O(1/t^\star)$ with high probability using $\widetilde O((d+\log(1/\delta))/p)$ samples.
\end{theorem}
The angular scale $O(1/t^\star)$ is exactly the scale needed for excess risk in the critical regime. To estimate the threshold, we use the threshold ERM of Algorithm~\ref{alg:threshold-erm} with the following guarantee:
\begin{theorem}[Informal: one-dimensional threshold ERM]
\label{thm:informal-threshold-erm}
For any fixed direction $\widehat w$, Algorithm~\ref{alg:threshold-erm} uses
$\widetilde O(1/\epsilon^2)$ independent validation samples and returns a threshold $\widehat t$ whose risk is within $\epsilon$ of the best threshold along $\widehat w$.
\end{theorem}
This leads us to the standalone sample-optimal halfspace learning guarantee.
\begin{theorem}[Informal: standalone halfspace learning]
\label{thm:informal-standalone-rv}
In the critical regime $p=\Theta(\epsilon)$, Algorithm~\ref{alg:rare-tail-direction} followed by Algorithm~\ref{alg:threshold-erm} returns a halfspace $\widehat f$ with excess risk $\le \epsilon$, with probability at least $1-\delta$, using $\widetilde O((d+\log(1/\delta))/\epsilon)+\widetilde O(1/\epsilon^2)$ samples, and runs in time $\widetilde O(dN)$.
\end{theorem}
In high-dimensions, the dimension-free threshold-validation term is lower order, so the sample complexity is governed by the rare-tail direction-estimation term. As a concrete specialization of the general theory, we study the probit noise profile. The resulting learning guarantee is in Corollary~\ref{cor:probit-excess-risk}.
\paragraph{Application to curriculum learning for RCN.}
The guarantees above do not cover RCN, since its noise is flat. Next, we use the rare-tail estimator as the warm start of the two-stage curriculum described earlier, to circumvent the statistical-query lower bound of \citet{diakonikolas2023nearoptimalboundslearninggaussian} for affine Gaussian halfspaces under RCN.
\begin{theorem}[Informal: curriculum learning for an RCN target]
\label{thm:informal-curriculum-rcn}
For $p = \Omega_\eta(\epsilon)$, there is a curriculum learning algorithm that learns the target halfspace to excess error $\epsilon$ using $\widetilde O(d/p)$ source samples and $\widetilde O_\eta((d/\epsilon) + 1/\epsilon^2)$ target RCN samples, with high probability.
\end{theorem}
In high dimensions, the leading target cost thus matches the information-theoretic rate, whereas the best known target-only learner \citep{diakonikolas2023nearoptimalboundslearninggaussian} uses $\widetilde O(d/\varepsilon+d/\max(p,\varepsilon)^2)$ samples.%

\section{Technical Preliminaries}
\paragraph{Notation.}
For $d \in \mathbb{N}$, let $I_d$ denote the $d \times d$ identity.
For $u,v \in \mathbb{R}^d$, write $\langle u,v\rangle$ for the Euclidean inner product and $\|u\|_2$ for the Euclidean norm.
Let $S^{d-1} = \{w \in \mathbb{R}^d : \|w\|_2 = 1\}$.
For unit vectors $u,v \in S^{d-1}$, write $\angle(u,v) = \arccos(\langle u,v\rangle) \in [0,\pi]$. We write $\mathbf{1}\{\cdot\}$ for indicator functions. All logarithms are natural. We use $\tilde{O}(\cdot)$ and $\tilde{\Omega}(\cdot)$ to suppress polylogarithmic factors in the relevant parameters (typically $d,1/\epsilon,1/\delta,t$). Throughout, all statements involving $\PP(\cdot | X=x), \E[\cdot | X=x]$ hold $P_X$-almost surely. We write $\phi$ and $\Phi$ to denote the standard normal density and CDF. We also use $\lambda(t) = \nicefrac{\phi(t)}{\Phi(-t)}$ for the inverse Mills ratio. 
\paragraph{Gaussian halfspace model.}
Fix an unknown $w^\star \in S^{d-1}$ and threshold $t^\star \in \mathbb{R}$.
We draw $X \sim \mathcal{N}(0,I_d)$ and define the signed margin $m(x) = \langle w^\star, x\rangle + t^\star$ and the corresponding halfspace $f^\star(x) = \sign(m(x))$, 
where $\sign(u)=+1$ for $u \ge 0$ and $\sign(u)=-1$ otherwise.

\paragraph{Risk and excess risk.}
For a distribution $\mathcal D$ over $\mathbb R^d\times\{\pm1\}$ and a classifier  $f:\mathbb R^d\to\{\pm1\}$, define the $0$--$1$ risk
\begin{align*}
    \operatorname{err}_{\mathcal D}(f)
    =
    \PP_{(X,Y)\sim\mathcal D}(f(X)\ne Y)
\end{align*}
Given a reference classifier $f^\star$, the excess risk of $f$ with respect to $f^\star$ is $\operatorname{err}_{\mathcal D}(f)-\operatorname{err}_{\mathcal D}(f^\star)$. When the distribution is clear from context, we write simply $\operatorname{err}(f)$. In the high bias regime, the following directional guarantee suffices for an $\epsilon$ excess risk guarantee.  The proof is deferred to Appendix~\ref{app:angle-to-risk}. 
\begin{lemma}[Angular error controls excess risk]
\label{lem:angle-to-risk-fixed-threshold}
In the Gaussian halfspace model described above, fix $t^\star \ge 1$, and let $w\in S^{d-1}$ be arbitrary, with induced halfspace $f_w(x) = \sign(\langle w, x\rangle +t^\star)$. There exist universal constants $c_0, C_0 > 0$ such that, whenever $\theta=\angle(w,w^\star)\le \nicefrac{c_0}{t^\star}$, we have
\begin{align*}
    \PP(f_w(X)\ne f^\star(X))
    \le
    C_0\theta\phi(t^\star)
\end{align*}
In particular, in the critical high-bias regime $p=\Theta(\epsilon)$, there is a sufficiently small universal constant $c>0$ such that $\angle(w,w^\star)\le \nicefrac{c}{t^\star}$ implies $\operatorname{err}(f_w)-\operatorname{err}(f^\star)\le \epsilon$
\end{lemma}

\subsection{Noise Families}
\label{sec:prelims:noise}
\paragraph{General margin-dependent label noise.}Let $g:[0,\infty)\to[0, \nicefrac{1}{2}]$ be measurable. We say $(X,Y)$ follows \emph{margin-dependent noise with profile $g$} if
\begin{align*}
\PP\big(Y \ne f^\star(X)\mid X=x\big) = g\big(|m(x)|\big)
\end{align*}
If $g$ is nonincreasing, we call this \emph{margin-monotone noise}. For a profile $g$, we write $D^g$ for the induced distribution over $(X,Y)$ under the current target halfspace; in particular, $D^\eta$ denotes the $\eta$-RCN distribution, where $g(u) = \eta ,\, \forall \, u\in \R$, and $\eta\in[0,\nicefrac{1}{2})$.%
\paragraph{Flat noise profiles.} Throughout this paper, when we refer to a noise profile as \emph{flat}, we mean RCN-like flat profiles where $g(\cdot)$ is a positive constant. The degenerate noiseless profile $g(\cdot) = 0$ is a flat function, but it does not create the obstruction studied here: it introduces no false negatives, and the rare-tail mean estimator achieves its strongest signal in precisely this case.
\paragraph{Probit noise.}\label{def:probit}
Fix $\sigma > 0$ and let $G \sim \mathcal{N}(0,1)$ be independent of $X$. In the probit model,
\begin{align*}
Y = \sign(\langle w^\star, \, X\rangle + t^\star + \sigma G)
\end{align*}
This is margin-monotone with $g(u)=\Phi(-u/\sigma)$ (with the convention $\Phi(-u/0)=0$ for $u>0$). 
\begin{definition}[Regularly varying margin noise]
\label{def:regularly-varying-margin-noise}
Fix constants $c>0$ and $C\ge 0$. A margin-monotone noise function $g$ has regular margin variation if $g$ is absolutely continuous, $g(u)\to 0$ as $u\to\infty$,  and for almost every $u\ge 1$ with $g(u)>0$,
\begin{align*}
    -g'(u)\ge\left(\frac{c}{(1+u)(\log(e+u))^{C}}\, \right) g(u)
\end{align*}
\end{definition}
This condition controls the relative decay rate of the noise profile, while placing no restrictions on the noise level near the boundary, where $g(0)$ may be arbitrarily close, or even equal, to $\nicefrac{1}{2}$. It says that the flip probability cannot remain essentially flat over large margin scales. On a multiplicative margin scale, such as from $u$ to $2u$, it forces an inverse-polylogarithmic decrease in the current value of $g$. This excludes constant RCN and long flat plateaus, but admits very slowly decaying noise functions, including logarithmic and sub-polynomial tails. Polynomial, exponential, logistic, probit, and sub-Gaussian noise all satisfy the condition.

\section{Overview of Techniques}
To establish the core ideas behind our main results, we first formalize the intuition that, in the biased regime, the typical samples are not the most informative ones. It is perhaps most apparent in the absence of noise, where $Y=f^\star(X)$. For $t^\star>0$, the rare-tail event corresponds to the negative side of the halfspace. When $0\le t^\star\le1$, this tail has constant mass, so Algorithm~\ref{alg:rare-tail-direction} only needs $\widetilde O(d+\log(1/\delta))$ samples. Our results are therefore stated for nontrivial $t^\star > 1$. %
\paragraph{Noiseless benchmark. } %
The relevant geometry becomes visible in two dimensions. When $t^\star$ is large, the separator is shifted away from the origin, so the Gaussian mass is heavily concentrated on one side of the boundary. Thus, a typical sample lies deep in the majority region and carries only weak directional information about the orientation of the halfspace. In contrast, the minority samples come from the thin tail cut off by the separator, and are therefore much more tightly coupled to the true normal.

Figure~\ref{fig:directional-overview} illustrates this distinction. Both panels show the same set of samples from a two-dimensional Gaussian with the same separator. In (a), the empirical Average vector induces a normal whose angle with $w^\star$ is relatively large, and hence the corresponding separator has a noticeable directional error. In (b), we instead use the direction induced by the rare-tail mean, namely, the normalized vector $-\widehat{\mu}_-$.  Its angle to $w^\star$ is smaller, as a consequence of Gaussian structure. In $d$ dimensions, we repeat the earlier decomposition $X=Z\,w^\star+X_\perp$. In the absence of noise, the minority event is exactly $\{Z<-t^\star\}$, and conditioning on this event only affects the coordinate $Z = \langle w^\star, x\rangle$, with $E[X\mid Y=-1]=E[Z\mid Z<-t^\star]w^\star=-\bigl(\nicefrac{\phi(t^\star)}{\Phi(-t^\star)}\bigr)w^\star$, and the population minority mean is therefore collinear with $w^\star$. Moreover, the norm is the inverse Mills ratio, which is of the order $t^\star$ in the large bias regime, so conditioning on the rare tail event actually amplifies the signal in the $w^\star$ direction by a factor of $\Theta(t^\star)$. 

The Average estimator enjoys the same symmetry, but not the same signal strength. Indeed, if we write $\mu_{\mathrm{avg}}=E[YX]$, then $\mu_{\mathrm{avg}}=-2E[X\mathbf{1}\{Y=-1\}]=-2\PP(Y=-1)E[X\mid Y=-1]=-2p\mu_-$, i.e., the Average vector is smaller than the minority mean by a factor of $2p$. Both estimators incur orthogonal noise of typical size $\Theta(\sqrt{d/N})$ for Average and $\Theta(\sqrt{d/(Np)})$ for the minority mean. Consequently, at the same sample budget $N=\tilde O(d/p)$, the minority mean typically reaches $O(1/t^\star)$-scale angular error, while Average remains at the larger $ O(1/(t^\star\sqrt p))$ scale. 

As it turns out, what matters for this benchmark is not exact noiselessness, but that the observed rare class remains tail-like, i.e. $p_-=O(p)$. Under $\eta$-RCN, $p_-=\eta+(1-2\eta)p$, so this condition holds whenever $\eta=O(p)$. The rare-tail mean therefore retains the $\widetilde O(d/p)$ rate even, after exact linear separability is destroyed. This contrasts with separability-based methods, which rely on the absence of label flips. By Lemma~\ref{lem:angle-to-risk-fixed-threshold}, two Gaussian halfspaces, sharing the same threshold, with normals at angle $\theta$, disagree on at most $O(\theta\phi(t^\star))$ mass. Therefore, when $p=\Phi(-t^\star)=\Theta(\epsilon)$, taking $\theta=O(1/t^\star)$ gives $O(p)=O(\epsilon)$ disagreement. Since excess risk under our noise models is bounded by disagreement with the Bayes halfspace, a correct threshold along such a direction suffices for $\epsilon$ excess risk.

\paragraph{Robustness beyond $\eta = O(p)$. }\label{subsec:noisy_geometry}
The signal amplification in the benchmark above extends to a broad and classical family of \emph{margin-dependent} corruption models, in which, the label flip probability depends only on the distance to the decision boundary. Consider the regularly-varying margin noise profiles of Definition~\ref{def:regularly-varying-margin-noise}. Under this weak condition, the same one-pass estimator achieves the $O(1/t^\star)$-scale angular error with the same sample complexity as in the noiseless regime.

Before stating the result, we note a small notational mismatch between the algorithm and the analysis. The proof uses the convention $t^\star>0$ and is written for the conditional mean $\mu_-=\mathbb E[X\mid Y=-1]$, while Algorithm~\ref{alg:rare-tail-direction} first identifies the empirical rare label and then orients the corresponding conditional mean. In practice, this is how one would go about implementing this estimator, since the rare label cannot be known apriori, but this difference in notation does not affect the stated sample-complexity bounds. In particular, if $p_-=\PP(Y=-1)$ is small, the empirical rare label is $-1$ with high probability, but if $p_-$ is instead bounded below by a constant, either label-conditional mean gives the same population direction up to a constant-factor in the sample complexity, since $\mathbb E[X]=0$ implies $p_-\mu_-+p_+\mu_+=0$. The formal reconciliation is discussed in Appendix~\ref{app:rare-label-reconciliation}.

\begin{theorem}[Direction recovery under regular variation]
\label{thm:rare-tail-direction-rv}
Assume $t^\star\ge1$,  and let $p=\Phi(-t^\star), p_- = \PP(Y=-1)$. If the labels follow a $(c,C)$-regularly varying margin-monotone noise profile then there is a sufficiently small universal constant $c_{\rm dir}>0$ such that Algorithm~\ref{alg:rare-tail-direction} returns a direction $\widehat w$ satisfying $\angle(\widehat w,w^\star)\le \nicefrac{c_{\rm dir}}{t^\star}$ with probability at least $1-\delta$, using $N_{dir}$ samples, where
\begin{align*}
    N_{\rm dir}
    =
    \widetilde O_{c,C}\left(
        (d+\log(1/\delta))
        \max\left\{
            \frac1p,\,
            \frac{(t^{\star})^{4}(\log(e+t^\star))^{2C}}{p_-}
        \right\}
    \right)
\end{align*}
\end{theorem}

\begin{proof}[Proof sketch]
Since this result is for direction recovery, we set $t=t^\star$ for readability. Let $\tau$ denote the target angle. Since the label depends on $X$ only through the margin coordinate $Z+t$, the population conditional mean $\mu_-$ is of the form
\begin{align*}
    \mu_-=\mathbb E[X\mid Y=-1]
    =
    \mathbb E[Z\mid Y=-1]w^\star,
\end{align*}
and is therefore aligned with the target $w^\star$. It remains to lower bound the norm of this conditional mean for empirical concentration. In Theorem~\ref{thm:minority_mean_general}, we show that the empirical rare-tail mean is within angle $\tau$ of $w^\star$ once there are $\widetilde O((d+\log(1/\delta))/(\|\mu_-\|_2^2\sin^2(\tau/2)))$ observed negatives, where the logarithmic factor comes from conditioning on a rare label. Since $N_-\ge p_-N/2$ with high probability and $\tau=c_{\rm dir}/t$, it suffices that
\begin{align}
    N=\widetilde O\left(\frac{t^2(d+\log(1/\delta))}{p_-\|\mu_-\|_2^2}\right)
    \label{eq:sample_bound_general}
\end{align}
Let $q(z)=\mathbb P(Y=-1\mid Z=z)$ and $U=|Z+t|$. Since $g$ is nonincreasing and $g\le1/2$, $q$ is nonincreasing, with a downward jump of $1-2g(0)$ at $z=-t$ and derivative $q'(z) = g'(|z+t|)$ for almost every $z\ne -t$. Gaussian integration by parts applied to $p_-\|\mu_-\|_2=-\mathbb E[Zq(Z)]$ then gives the following signal identity that captures the intuition as to why non-flat profiles amplify the signal
\begin{align}
    p_-\|\mu_-\|_2=(1-2g(0))\phi(t)+\mathbb E[-g'(U)]. 
    \label{eq:signal-identity}
\end{align}
Notice that the first term is due to the jump of $q$ at the boundary, which is the only part of the signal that survives in flat profiles, and is at most of order $tp$. The second term, involving the expected derivative, captures signal variation away from the boundary. This additional signal is precisely the edge that boundary-local profiles have over flat profiles. To facilitate a closed-form analysis of this term, we will use the regular variation condition from Definition~\ref{def:regularly-varying-margin-noise}, although eq~\eqref{eq:signal-identity} is more general and applies to all absolutely continuous margin-monotone profiles for this problem. 

The observed negatives come from two groups, namely samples from the clean tail $\{Z<-t\}$, and false negatives from $\{Z\ge-t\}$. Since $g\le1/2$, every clean sample keeps its label $-1$ with probability at least $1/2$, so the first group alone has probability at least $p/2$, and $p_-\ge p/2$ always. 

We first consider the tail-like regime $p_-\le2p$, where the observed negatives have a distribution closer to the true negatives. The first group then makes up at least a $\nicefrac{p}{2p_-}\ge1/4$ fraction of the observed negatives, and all of these lie at $Z<-t$. The false negatives make up the rest, but because $g$ is nonincreasing, flips are more likely close to the boundary, and false negatives have mean at most $\mathbb E[Z\mid Z\ge-t]\le3/10$. Together, this gives $\|\mu_-\|_2=-\mathbb E[Z\mid Y=-1]\ge t/4-9/40\ge t/40$, and since $p_-\ge p/2$, applying \eqref{eq:sample_bound_general} gives the first term of $N_{\rm dir}$. For this tail-like regime, regular variation is unnecessary.

The main technical difficulty is in the inflated regime $p_->2p$, where false negatives make up more than half of the observed negatives. They increase $p_-$ without adding to the boundary-jump term, which on its own only gives $\|\mu_-\|_2\ge(1-2g(0))\phi(t)/p_-$, of order $(1-2g(0))\,tp/p_-$. This degrades as $p_-/p$ grows, as in $\eta$-RCN where $p_- \approx \eta$, and vanishes when $g(0)=1/2$, as for probit noise. The boundary-jump term alone need not provide enough signal, and we shift focus to the variation term instead. For a suitable constant $L$, Definition~\ref{def:regularly-varying-margin-noise} gives $-g'(u)\ge c_1g(u)/(t(\log(e+t))^C)$ for almost every $u\in[1,Lt]$, where $c_1=c_1(c,C)>0$. In other words, on margins up to order $t$, the flip probability decays at a relative rate of at least $1/t$, up to logarithmic factors. Since the condition only constrains margins $u\ge1$, we split according to where these flips happen. 

Suppose first that a constant fraction of the flips happen away from the boundary, say $\mathbb E[g(U)\mathbf1\{U\ge1\}]\ge p_-/4$. For a suitable universal constant $L$, $\mathbb P(U>Lt)\le p/2$. Since $g\le1/2$,
\begin{align*}
    \mathbb E[g(U)\mathbf1\{U>Lt\}]
    \le
    \frac12\mathbb P(U>Lt)
    \le
    \frac p4
    <
    \frac{p_-}{8}
\end{align*}
and at least half of these flips lie in the window $[1,Lt]$. Consequently,
\begin{align*}
    \mathbb E[-g'(U)]
    \ge
    \frac{
        c_1\mathbb E[g(U)\mathbf1\{1\le U\le Lt\}]
    }{t(\log(e+t))^C}
    \ge
    \frac{c_1p_-}{8t(\log(e+t))^C}
\end{align*}
Since the boundary-jump term is nonnegative, eq.~\eqref{eq:signal-identity} gives $\|\mu_-\|_2=\Omega_{c,C}(1/(t(\log(e+t))^C))$, and applying \eqref{eq:sample_bound_general} gives the second term of $N_{\rm dir}$.

The other possibility is that $\mathbb E[g(U)\mathbf1\{U\ge1\}]<p_-/4$, so most flips happen within distance $1$ of the boundary. The observed negatives with $Z>1-t$ are flips at margin larger than $1$, so they have probability less than $p_-/4$, and at least three quarters of the observed negatives lie at
$Z\le1-t\le0$. Set $b=1-t$. On each side of $b$, conditioning on $Y=-1$ reweights the truncated Gaussian by the nonincreasing function $q$. 
Then the two conditional means are at most $\mathbb E[Z\mid Z\le b]\le-\sqrt{2/\pi}$ and $\mathbb E[Z\mid Z>b]\le\sqrt{2/\pi}$, respectively. Combining their contributions gives $\mathbb E[Z\mid Y=-1]\le -\nicefrac12\sqrt{\nicefrac2\pi}$. Therefore, $\|\mu_-\|_2=\Omega(1)$, and \eqref{eq:sample_bound_general} requires only $\widetilde O((d+\log(1/\delta))t^2/p_-)$ samples, which is weaker than the second term of $N_{\rm dir}$.

Combining the three cases proves the theorem. Finally, since $p_-\ge p/2$ and $t^2\le2\log(1/p)$, the bound simplifies to $\widetilde O_{c,C}((d+\log(1/\delta))/p)$ after suppressing polylogarithmic factors in $1/p$, which then matches the noiseless rate.
\end{proof}
\paragraph{Threshold estimation.}
Once the direction is accurate, estimating the threshold is a one-dimensional problem. We use Algorithm~\ref{alg:threshold-erm}, which projects the validation samples onto the recovered direction, sorts the projected values, and scans the induced threshold intervals.
\begin{theorem}[Threshold ERM along a fixed direction]
\label{thm:threshold-erm-main}
Fix a unit vector $\widehat w$ independent of a validation sample of size $k$, and let
$\widehat t$ be the threshold returned by Algorithm~\ref{alg:threshold-erm}. For any
$\epsilon,\delta\in(0,1)$, with
$k=\widetilde O(\epsilon^{-2}\log(1/\delta))$,  
\begin{align*}
\operatorname{err}\!\left(\operatorname{sign}(\langle \widehat w,X\rangle+\widehat t)\right)
    \le
    \inf_{b\in\mathbb R}
    \operatorname{err}\!\left(\operatorname{sign}(\langle \widehat w,X\rangle+b)\right)
    +
    \epsilon
\end{align*} with probability at least $1-\delta$.
Moreover, Algorithm~\ref{alg:threshold-erm} runs in time $O(dk+k\log k)$.
\end{theorem}
The proof uses the standard VC bound for thresholds on the real line. For the final learner, we apply Theorems~\ref{thm:threshold-erm-main} and ~\ref{thm:rare-tail-direction-rv} with accuracy parameter set to a sufficiently small constant multiple of the target $\epsilon$, and adjust the failure probability via a union bound. The full proof is in Appendix~\ref{app:subsec:threshold-erm}. 
\begin{theorem}[Halfspace learning under regular variation]
\label{thm:standalone-rv-halfspace}
Assume the setting of Theorem~\ref{thm:rare-tail-direction-rv}, and suppose
$p=\Phi(-t^\star)=\Theta(\epsilon)$. There is an algorithm that, given
\begin{align*}
    N
    =
    \widetilde O_{c,C}\left(
        (d+\log(1/\delta))
        \max\left\{
            \frac1p,\,
            \frac{(t^\star)^{4}(\log(e+t^\star))^{2C}}{p_-}
        \right\}
        +
        \frac1{\epsilon^2}
    \right)
\end{align*}
independent labeled samples, returns a halfspace $\widehat f = (\widehat w, \widehat t)$ such that $\operatorname{err}(\widehat f)-\operatorname{err}(f^\star)\le \epsilon$ with probability at least $1-\delta$. The algorithm runs in time $\widetilde O(dN)$.
\end{theorem}
\paragraph{A canonical specialization: probit noise.} The probit model is a special case of regularly-varying margin-dependent noise profiles, and the following learning guarantee is almost immediate from the proof of Theorem~\ref{thm:rare-tail-direction-rv}. The proof is deferred to Appendix~\ref{app:probit}. 
\begin{corollary}[Excess-risk guarantee under probit noise]
\label{cor:probit-excess-risk}%
Assume the probit model with $\sigma>0$ and $t^\star\ge1$. Let $p=\Theta(\epsilon),p_\sigma=\Phi\left(-\nicefrac{t^\star}{\sqrt{1+\sigma^2}}\right)$. There is an algorithm, that, given
\begin{align*}
    N
    =
    \widetilde O\left(
        \frac{(d+\log(1/\delta))(1+\sigma^2)^2}{p_\sigma}
        +
        \frac{1}{\epsilon^2}
    \right)
\end{align*}
independent probit samples, returns a halfspace $\widehat f$ such that $\operatorname{err}(\widehat f)-\operatorname{err}(f^\star)\le\epsilon$ with probability at least $1-\delta$. The algorithm runs in time $\widetilde O(dN)$.
\end{corollary}

\section{Applications to Curriculum Learning}\label{sec:curriculum}
The guarantees of Section~\ref{subsec:noisy_geometry} need the noise to decay away from the boundary, so they do not cover flat profiles such as $\eta$-RCN~\citep{angluin1988learning}. For biased Gaussian halfspaces, this is also where efficient learning appears to be expensive, as both the best known efficient learner and the statistical-query lower bound of \citet{diakonikolas2023nearoptimalboundslearninggaussian} depend quadratically on $1/p$. We first identify where this cost comes from, and then show that one additional stage of samples with boundary-local noise removes it.

\paragraph{The cost.} Under $\eta$-RCN with $t^\star>0$, the label $-1$ is observed with probability $1-\eta$ on the tail side of the separator and $\eta$ on the other, so the label distribution jumps at the separator and is constant elsewhere. A first-moment statistic, such as the rare-tail mean or the Average vector $\mathbb{E}[YX]$, therefore sees the direction of $w^\star$ only through this jump, weighted by the Gaussian density at the separator, $\phi(t^\star)\approx t^\star p$. This is the boundary-jump term $(1-2\eta)\phi(t^\star)$ of eq.~\eqref{eq:signal-identity}, which is all that survives under flat noise, and since it is of order $t^\star p$, such statistics need on the order of $d/p^2$ samples to reach angle $O(1/t^\star)$. The best known efficient learner pays the same price in its warm start. \citet{diakonikolas2023nearoptimalboundslearninggaussian} initialize their algorithm with the Average vector, and this accounts for the only quadratic term in their sample complexity $\widetilde O(d/\varepsilon+d/\max\{p,\varepsilon\}^2)$, since once a direction within angle $1/(5t^\star)$ is available, the remaining steps use the information-theoretically optimal $\widetilde O(d/\varepsilon)$ samples~\citep{massart2006}. Their lower bound suggests that efficient methods cannot avoid the quadratic dependence on $1/p$ with RCN samples alone. Efficient statistical-query (SQ) algorithms~\citep{kearns1998efficient}, which access the data only through averages of bounded functions, as Algorithm~\ref{alg:rare-tail-direction} and gradient methods do, need $\widetilde\Omega(\sqrt d/\max\{p,\varepsilon\}^2)$ samples, and so do low-degree polynomial tests. Since the later steps are cheap, whenever this lower bound exceeds $d/\varepsilon$, it is in effect a lower bound on finding the warm start from RCN samples.

\paragraph{A simple curriculum.} If no average of RCN samples reveals the direction at the information-theoretic rate, we can instead compute the same average on samples from a slightly different distribution. \citet{cornacchia2023mathematical} prove an analogous separation for parities, where a curriculum that starts from a biased product distribution lets a two-layer network learn $k$-parities at $\mathrm{poly}(d)$ cost, while noisy-gradient training on the uniform distribution alone requires $d^{\Omega(k)}$ samples. Their curriculum changes the input distribution between the stages, but ours does not need to. Our source stage keeps the Gaussian marginal and the Bayes halfspace, and differs from the target only in its noise profile, which needs to decay away from the boundary (Definition~\ref{def:regularly-varying-margin-noise}). This is the least intrusive way to add information, since it requires no control over which instances are labeled, only over who labels them. Careful human annotators, whose mistakes concentrate on ambiguous instances~\citep{whitehill2009whose,beigman2009}, are a natural source of such labels.

\begin{definition}[Two-stage curriculum learning for Gaussian halfspaces]
\label{def:two-stage-curriculum}
Fix stage lengths $(n_1,n_2)$, a source profile $g_1:[0,\infty)\to[0,1/2]$, and a target noise rate $\eta<1/2$. Nature fixes an unknown target halfspace $f^\star=f_{w^\star,t^\star}$, and defines
\begin{align*}
    D_1=D^{g_1}_{w^\star,t^\star},
    \qquad
    D_2=D^\eta_{w^\star,t^\star}
\end{align*}
Both stages share the same Gaussian marginal and Bayes halfspace, but have different label noise profiles. The learner first receives source samples $S_1\sim D_1^{n_1}$ and computes an intermediate state $M_1$, such as an initialization, before the target samples are revealed. It then receives independent target samples $S_2\sim D_2^{n_2}$ and, using $M_1$ and $S_2$, outputs a classifier $\widehat f:\mathbb R^d\to\{\pm1\}$. The target-stage procedure accesses the source samples only through $M_1$. The goal is to ensure the excess risk with respect to $D_2$ satisfies $\err_2(\widehat f)-\err_2(f^\star)\le \epsilon$.
\end{definition}

\begin{table}[t]
\centering
\small
\setlength{\tabcolsep}{2.6pt}
\renewcommand{\arraystretch}{1.22}
\begin{minipage}[t]{0.495\linewidth}
\centering
\textbf{Sample complexity}
\vspace{0.35em}
\begin{tabular}{@{}lcc@{}}
\toprule
Bias regime
& This paper
& \citet{diakonikolas2023nearoptimalboundslearninggaussian}
\\
\midrule
$p<\epsilon$
&
$O(1)$
&
$O(1)$
\\

$p=\Theta(\epsilon)$
&
\textcolor{blue!65!black}{$\widetilde O(d/\epsilon)$}
&
\textcolor{red!70!black}{$\widetilde O(d/\epsilon^2)$}
\\

$\epsilon \ll p\le \sqrt{\epsilon}$
&
\textcolor{blue!65!black}{$\widetilde O(d/\epsilon)$}
&
\textcolor{red!70!black}{$\widetilde O(d/\epsilon+d/p^2)$}
\\

$p>\sqrt{\epsilon}$
&
$\widetilde O(d/\epsilon)$
&
$\widetilde O(d/\epsilon)$
\\
\bottomrule
\end{tabular}
\end{minipage}
\hfill
\begin{minipage}[t]{0.495\linewidth}
\centering
\textbf{Runtime}
\vspace{0.35em}
\begin{tabular}{@{}lcc@{}}
\toprule
Bias regime
& This paper
& \citet{diakonikolas2023nearoptimalboundslearninggaussian}
\\
\midrule
$p<\epsilon$
&
$O(1)$
&
$O(1)$
\\

$p=\Theta(\epsilon)$
&
\textcolor{blue!65!black}{$\widetilde O(d^2/\epsilon)$}
&
\textcolor{red!70!black}{$\widetilde O(d^2/\epsilon^4)$}
\\

$\epsilon \ll p\le \sqrt{\epsilon}$
&
\textcolor{blue!65!black}{$\widetilde O(d^2/\epsilon)$}
&
\textcolor{red!70!black}{$\widetilde O(d^2/\epsilon^3+d^2/(p^2\epsilon^2))$}
\\

$p>\sqrt{\epsilon}$
&
\textcolor{blue!65!black}{$\widetilde O(d^2/\epsilon)$}
&
\textcolor{red!70!black}{$\widetilde O(d^2/\epsilon^3)$}
\\
\bottomrule
\end{tabular}
\end{minipage}
\caption{Bias-regime comparison for learning an $\eta$-RCN target. We suppress polylogarithmic factors in $d,1/\epsilon,1/\delta,t$ and constants depending on fixed $\eta<1/2$, and display only leading high-dimensional terms. Our curriculum learner uses source and target samples, while \citet{diakonikolas2023nearoptimalboundslearninggaussian} use target samples only. For $p<\epsilon$, a constant classifier suffices, and in the nontrivial regimes, the information-theoretic target sample rate is $\widetilde\Theta(d/\epsilon)$~\citep{massart2006}.}
\label{tab:bias-regime-comparison}
\end{table}
\begin{theorem}[Two-stage curriculum guarantee for an $\eta$-RCN target]
\label{thm:curriculum-rcn-target}
Assume the two-stage model of Definition~\ref{def:two-stage-curriculum}. The source stage satisfies
the assumptions of Theorem~\ref{thm:rare-tail-direction-rv} with $p_-^{(1)} = \PP_{D_1}(Y=-1)$, and the target stage is $\eta$-RCN
with known constant $\eta<1/2$. For $t^\star\ge1$ and $p=\Phi(-t^\star)\ge C_\eta\epsilon$, if
\begin{align*}
    n_1
    =
    \widetilde O_{c,C}\left(
        (d+\log(1/\delta))
        \max\left\{
            \frac1p,\,
            \frac{(t^\star)^{4}(\log(e+t^\star))^{2C}}{p_-^{(1)}}
        \right\}
    \right)
\end{align*}
source samples and
\begin{align*}
    n_2
    =
    \widetilde O_\eta\left(
        \frac{d+\log(1/\delta)}{\epsilon}
        +
        \frac{1}{\epsilon^2}
    \right)
\end{align*}
target RCN samples are used, then Algorithm~\ref{alg:rare-tail-curriculum-rcn} returns a halfspace
$\widehat f$ satisfying $\operatorname{err}_{D_2}(\widehat f)-\operatorname{err}_{D_2}(f^\star)\le\epsilon$ with probability at least $1-\delta$. The runtime is $\widetilde O_\eta(d(n_1+n_2))$.
\end{theorem}
\begin{remark}[Comparison with target-only learning]
Table~\ref{tab:bias-regime-comparison} compares Theorem~\ref{thm:curriculum-rcn-target} with learning from target samples alone. When $d\gtrsim 1/\varepsilon$, the curriculum learner uses $\widetilde O(d/\varepsilon)$ samples in every nontrivial bias regime, which is the information-theoretic rate. This does not contradict the lower bound of \citet{diakonikolas2023nearoptimalboundslearninggaussian}, since the warm start now comes from source samples, and with a flat source the quadratic cost would return. When $p>\sqrt\varepsilon$, the target-only algorithm is already sample-optimal, and the gain is in runtime. We defer proofs to Appendix~\ref{app:curriculum-learning}.
\end{remark}

\section{Limitations and Open Questions}
Like much of the algorithmic literature on learning halfspaces with label noise, our results assume Gaussian covariates. In particular, this is the setting of the $\eta$-RCN baseline and of its SQ lower bound~\citep{diakonikolas2023nearoptimalboundslearninggaussian}, and, to our knowledge, the only one in which affine halfspaces with label noise are known to be learnable to excess risk $\varepsilon$ at near-optimal rates in polynomial time (see Appendix~\ref{app:sec:related_work}). Nonetheless, our analysis is not particularly tied to Gaussians: the alignment of the minority mean with $w^\star$ only uses rotational symmetry about $w^\star$, and the signal identity of eq.~\eqref{eq:signal-identity} keeps its form for any such law, with $\phi$ replaced by $\Psi(z)=\int_z^\infty u\rho(u)\,du$, where $\rho$ is the density of $\langle w^\star,X\rangle$. The assumption appears in the tail and concentration bounds, and is a strict requirement only for the target-stage requirement of Section~\ref{sec:curriculum}, where it is inherited from \citep{diakonikolas2023nearoptimalboundslearninggaussian}. We therefore expect the direction-recovery guarantees to carry over to rotationally invariant laws with comparable tails. Rotational symmetry, however, is harder to remove. For an isotropic law, $\mathbb{E}[X\mid\langle w,X\rangle]$ is parallel to $w$ for every direction $w$ only if the law is spherically symmetric~\citep{eaton}. On other isotropic marginals, including log-concave ones, the minority mean can point away from $w^\star$, and the estimator would need some form of correction.
 
Consistent with the literature discussed in Appendix~\ref{app:sec:related_work}, our curriculum result assumes that the source samples arrive in a separate stage before the target RCN samples, and it does not cover the case when a single pool of samples is provided to the learner, but of mixed annotation quality. A pool with an $\alpha$ fraction of source samples has the margin-monotone profile $\alpha g_1+(1-\alpha)\eta$, and since the boundary-jump term of eq.~\eqref{eq:signal-identity} is at most of order $t^\star p$, Algorithm~\ref{alg:rare-tail-direction} run directly on the pool still gains from the variation term $\alpha\mathbb E[-g_1'(U)]$ whenever the source noise varies with the margin. Perhaps surprisingly, this would fail if the source samples were noiseless, since the pool is then RCN with the smaller rate $(1-\alpha)\eta$, and unless this rate is $O(p)$, the clean source samples have to be identified first. When each instance is labeled several times by known annotators, as is common in crowdsourcing, spammer scores that detect annotators who label without looking~\citep{raykar} give a simple way to do this, and designing such filters under minimal assumptions is a natural next step.

\begin{ack}
This work is supported in part by NSF award \# 2217058 (EnCORE Institute).

\end{ack}

\section*{Statement on AI Use}
All ideas and technical contributions in this paper are the authors' own. ChatGPT 5.5 Plus and, later, Opus 5.5 were used for language editing, proofreading, and transcribing handwritten proofs to \LaTeX. ChatGPT 5.5 Plus was also used to transcribe a hand-drawn version of Figure~\ref{fig:directional-overview} into TikZ code. The authors bear sole responsibility for the content and correctness of this manuscript.

\bibliographystyle{plainnat}
\bibliography{references}

\newpage
\appendix

\section{Technical appendices and supplementary material}

\subsection{Related Work}
\label{app:sec:related_work}

\paragraph{Noisy halfspace learning.}
Learning from noisy labels has been central to computational learning theory since \citet{angluin1988learning} introduced random classification noise (RCN). \citet{kearns1998efficient} then showed that any algorithm that accesses the data only through statistical queries (SQ), i.e., averages of bounded functions, can be simulated from RCN samples, and for halfspaces, this led to polynomial-time learners that make no assumption on the marginal distribution~\citep{blum1998polynomial,cohn1997learning}. Massart noise~\citep{massart2006} sits between RCN and adversarial noise. Here, the flip probability may depend on the instance arbitrarily, but can never exceed some $\eta<1/2$. For homogeneous halfspaces, error $\mathrm{opt}+\varepsilon$ can be reached in polynomial time under structured marginals, for instance under the uniform distribution on the unit ball, but only for very small $\eta<1.8\times 10^{-6}$~\citep{awasthi2015efficient}. Similar results exist for a broader family of marginals under Massart noise~\citep{diakonikolas2020massart_structured} and Tsybakov noise~\citep{tsybakov2004optimal,diakonikolas2021tsybakov}, which allows the flip probability to approach $1/2$ on a fraction of the instances. For affine halfspaces under Gaussian marginals, the algorithm of \citet{diakonikolas2022generalhalfspacesmassart} runs in time $d^{O(\log(1/\gamma))}\mathrm{poly}(1/\varepsilon)$, where $\gamma=\max\{p,\varepsilon\}$, which is quasi-polynomial when $p$ is of order $\varepsilon$. Margin-monotone noise with $g(0)<1/2$ is a special case of Massart noise, but since their algorithm is designed for the worst case over all Massart noise, the resulting guarantee is no better on these profiles. In contrast, for the regularly varying profiles of Definition~\ref{def:regularly-varying-margin-noise}, our learner runs in polynomial time with $\widetilde O(d/p)$ samples for the direction. In this sense, our result is a beyond-worst-case counterpart of theirs~\citep{balcan2021noise}.

\paragraph{Margin-dependent noise.}
Noise that depends on an instance only through its distance to the decision boundary, and does not increase with it, goes back to the monotonic noise of \citet{bylander1998learning}. \citet{Servedio1999OnPL}
showed that a very simple algorithm, which he called Averaging and which estimates the degree-one Chow parameters $\mathbb{E}[YX]$ by summing the examples multiplied by their labels, learns homogeneous halfspaces under this noise when the marginal is uniform on the unit sphere. Our rare-tail mean is a variant of this sum, with the labels centered at the majority label. The two behave similarly for homogeneous halfspaces, but with a bias $p$, the sum needs $\widetilde O(d/p^2)$ samples to reach the angular accuracy that the rare-tail mean reaches with $\widetilde O(d/p)$, even without noise. The testable learner of \citet{diakonikolas2024testable} also averages the minority-label points, but only to locate a point near the boundary, around which the problem becomes nearly homogeneous. Under Gaussian marginals and logistic noise, \citet{hsu2024sample} determine, up to logarithmic factors, the sample complexity of estimating the direction of an homogeneous halfspace, but they know of efficient estimators that achieve it only at high noise levels and without noise. In the biased regime, where an angle of order $1/t^\star$ suffices, the rare-tail mean is efficient for every regularly varying profile, including the logistic and probit ones. These have $g(0)=1/2$, so they are not Massart, and although they satisfy a Tsybakov condition, the polynomial-time Tsybakov results above are for homogeneous halfspaces, while ours have a bias.

\paragraph{The choice of marginal.}
For general halfspaces with label noise, the algorithms closest to ours are all stated for Gaussian marginals. They reach excess risk $\varepsilon$ under RCN~\citep{diakonikolas2023nearoptimalboundslearninggaussian}, run in quasi-polynomial time under Massart noise~\citep{diakonikolas2022generalhalfspacesmassart}, and reach error $O(\mathrm{opt}+\varepsilon)$ under adversarial noise~\citep{diakonikolas_nasty_noise_2018,diakonikolas_general_halfspace_adversarial_noise_2022}, which, under RCN in the biased regime, the constant classifier already does. Without assumptions on the marginal, learners pay for their generality. Under RCN, their sample complexity is polynomial in $d$, $1/\varepsilon$, $1/(1-2\eta)$ and the number $b$ of bits used to write each example~\citep{blum1998polynomial}, while $\widetilde O(d/\varepsilon)$ samples suffice information-theoretically, independently of $b$~\citep{massart2006}. Under Massart noise, they reach only error $\eta+\varepsilon$~\citep{diakonikolas2019distribution,ankur_evolvability}, which can be far from $\mathrm{opt}+\varepsilon$, a gap that efficient SQ algorithms cannot close~\citep{ankur_evolvability,diakonikolas2022nearoptimalmassart,nasser2022optimal}. Under log-concave marginals, polynomial regression reaches error $\mathrm{opt}+\varepsilon$ for any label noise, which covers the setting of this paper, but with running time and samples of the form $d^{f(1/\varepsilon)}$ for a growing function $f$~\citep{kalai2008agnostically}. With Massart noise, the only efficient learners we know of that reach excess risk $\varepsilon$ for general halfspaces beyond Gaussian marginals use comparison queries, a stronger oracle than labels alone~\citep{hopkins2020noise}. Since Gaussian distributions are log-concave, the SQ lower bounds below also apply to any SQ algorithm that succeeds for all log-concave marginals.

\paragraph{Lower bounds and data access.}
SQ lower bounds are the standard evidence that these problems are computationally hard, and they apply to every algorithm that, like Algorithm~\ref{alg:rare-tail-direction} and gradient methods, accesses the data only through averages. Under Gaussian marginals, SQ algorithms for general halfspaces need complexity $d^{\Omega(\log(1/\gamma))}$ under Massart noise~\citep{diakonikolas2022generalhalfspacesmassart}, matching the algorithm above. Under RCN, efficient SQ algorithms and low-degree polynomial tests need $\Omega(\sqrt d/\max\{p,\varepsilon\}^2)$ samples, with a quadratic dependence on $1/p$ that the information-theoretic rate does not have~\citep{diakonikolas2023nearoptimalboundslearninggaussian}. These lower bounds, however, concern the distribution that the queries are answered from, so they do not apply to learners with additional access. For general Gaussian halfspaces, \citet{diakonikolas2024active} show that label queries on a pool of unlabeled points do not help much unless the pool is exponentially large in $d$, while membership queries reduce the label complexity to $\widetilde O(\min\{1/p,1/\varepsilon\}+d\,\mathrm{polylog}(1/\varepsilon))$ for the weaker target $O(\mathrm{opt})+\varepsilon$. Membership queries, however, ask for labels of points that the learner constructs, and as \citet{awasthi2013local} note, querying such unnatural points increases the noise of human labelers. Our curriculum needs no such queries, only some labels from annotators whose mistakes concentrate near the boundary.

\paragraph{Curriculum learning.} 
Curriculum learning was popularized by \citet{bengio2009curriculum} as the principle that learning can benefit from examples being presented to a learner in a meaningful order. Curricula that move from reliable to noisy labels have been among the most effective strategies on noisy-label benchmarks~\citep{Guo_2018_ECCV,mentor_net}, and a controlled study finds that curricula help primarily when the labels are noisy or training budget is limited~\citep{wu2021curricula}. Theoretical models of curriculum learning are fewer. For learning parities under the uniform distribution, which is another problem where there is a proven gap for SQ methods, they often change the input distribution between the stages. \citet{cornacchia2023mathematical}, for instance, move from a biased product distribution to the uniform one, and \citet{abbe2023curriculum} from sparse inputs to dense ones. Moreover, their mathematical modeling of curriculum learning is specific to the training dynamics of a neural network, through SGD or noisy gradient methods. Our formulation is instead algorithm-agnostic. In Appendix~\ref{subsec:gen-cl-model}, we give a general definition of curriculum learning for an arbitrary
binary hypothesis class and any finite number of stages, and then specialize to the two-stage model described in Section~\ref{sec:curriculum} for noisy Gaussian halfspaces.

\section{Learning Under Regular Variation Margin Monotone Profiles}

Throughout, $X\sim\mathcal{N}(0,I_d)$, the target halfspace is
$f^\star(x)=\sign(\langle w^\star,x\rangle+t)$ with $w^\star\in\mathbb{S}^{d-1}$ and $t>0$, and we write
$Z=\langle w^\star,X\rangle\sim\mathcal{N}(0,1)$ and $M=Z+t\sim\mathcal{N}(t,1)$.
We denote the standard normal density and CDF by
\begin{align*}
\phi(u) = (2\pi)^{-1/2}e^{-u^2/2}, \qquad
\Phi(u) = \int_{-\infty}^u \phi(s)\,ds
\end{align*}
and write the halfspace bias as $p=\PP(f^\star(X)=-1)=\Phi(-t)$ and the inverse Mills ratio as $\lambda(t)=\phi(t)/\Phi(-t)$.
\subsection{Gaussian identities}

\begin{lemma}[One-dimensional decomposition]\label{lem:gauss_decomposition}
Let $X\sim\cN(0,I_d)$ and $w^\star\in\mathbb{S}^{d-1}$
Define $Z=\langle w^\star,X\rangle$ and $X_\perp=X-Zw^\star$
Then $Z\sim\cN(0,1)$, $X_\perp\sim \cN(0,I_d-w^\star w^{\star\top})$, and $Z\perp X_\perp$ 
Consequently, for any integrable $\psi:\R\to\R$,
\begin{align*}
\E[X\,\psi(Z)] = w^\star\,\E[Z\,\psi(Z)]
\end{align*}
\end{lemma}

\begin{proof}
Rotational invariance gives that $(Z,X_\perp)$ is jointly Gaussian with $\E[Z]=0$, $\E[X_\perp]=0$,
$\Var(Z)=1$, $\Cov(Z,X_\perp)=0$, and $\Cov(X_\perp)=I_d-w^\star w^{\star\top}$
Hence $Z\perp X_\perp$ and $X=Zw^\star+X_\perp$
The displayed identity follows by conditioning on $Z$ and using $\E[X_\perp\mid Z]=0$
\end{proof}

\begin{lemma}[Truncation and Mills bounds]\label{lem:mills}
For every $t>0$,
\begin{align*}
\E[Z\,\mathbf{1}\{Z<-t\}] = -\phi(t),
\qquad
\E[Z\mid Z<-t] = -\lambda(t)
\end{align*}
Moreover,
\begin{align*}
\frac{t}{t^2+1}\,\phi(t) \le \Phi(-t) \le \frac{\phi(t)}{t},
\qquad
t \le \lambda(t) \le t + \frac{1}{t}
\end{align*}
\end{lemma}

\begin{proof}
Integrating by parts,
\begin{align*}
\E[Z\,\mathbf{1}\{Z<-t\}] = \int_{-\infty}^{-t} z\phi(z)\,dz
= \Big[-\phi(z)\Big]_{-\infty}^{-t} = -\phi(t)
\end{align*}
so $\E[Z\mid Z<-t]=-\phi(t)/\Phi(-t)=-\lambda(t)$
The Mills bounds are the standard integration-by-parts inequalities 
\end{proof}

\subsection{From fixed-threshold angular error to excess risk}
\label{app:angle-to-risk}
\begin{lemma}[Restatement of Lemma~\ref{lem:angle-to-risk-fixed-threshold}]
In the Gaussian halfspace model described above, fix $t \ge 1$, and let $w\in S^{d-1}$ be arbitrary, with induced halfspace $f_w(x) = \sign(\langle w, x\rangle +t)$. There exist universal constants $c_0, C_0 > 0$ such that, whenever $\theta=\angle(w,w^\star)\le \frac{c_0}{t}$, we have
\begin{align*}
    \PP(f_w(X)\ne f^\star(X))
    \le
    C_0\theta\phi(t)
\end{align*}
In particular, in the critical high-bias regime $p=\Theta(\epsilon)$, there is a sufficiently small universal constant $c>0$ such that $\angle(w,w^\star)\le \frac{c}{t}$ implies
\begin{align*}
    \operatorname{err}(f_w)-\operatorname{err}(f^\star)\le \epsilon
\end{align*}
\end{lemma}
\begin{proof}
We write $f_w(x)=\operatorname{sign}(\langle w,x\rangle+t)$. If $d=1$, the angle hypothesis forces $\theta=0$ after taking $c_0<\pi$, and the claim is immediate. We will therefore assume $d\ge2$. 

By rotational invariance, we may assume, without loss of generality, 
\begin{align*}
    w^\star=e_1,
    \qquad
    w=\cos\theta\,e_1+\sin\theta\,e_2
\end{align*}
Let $G_1,G_2$ be independent standard Gaussians. Then
\begin{align*}
    \langle w^\star,X\rangle+t
    =
    G_1+t
\end{align*}
and
\begin{align*}
    \langle w,X\rangle+t
    =
    \cos\theta\,(G_1+t)+\sin\theta\,G_2+t(1-\cos\theta)
\end{align*}
If the two halfspaces disagree, these two signed margins have opposite signs. Since
$\cos\theta>0$ for $\theta\le c_0/t$ and $c_0$ sufficiently small, this implies
\begin{align*}
    |G_1+t|
    \le
    \frac{|\sin\theta|}{\cos\theta}|G_2|
    +
    \frac{t(1-\cos\theta)}{\cos\theta}
\end{align*}
For $\theta\le c_0/t$, after choosing $c_0$ sufficiently small, we have
\begin{align*}
    \frac{|\sin\theta|}{\cos\theta}\le 2\theta,
    \qquad
    \frac{t(1-\cos\theta)}{\cos\theta}\le 2t\theta^2
\end{align*}
Therefore
\begin{align*}
    \PP(f_w(X)\ne f^\star(X))
    \le
    \PP\left(|G_1+t|\le 2\theta |G_2|+2t\theta^2\right)
\end{align*}
Conditioning on $G_2$, and using the mean value theorem, for every $r\ge0$,
\begin{align*}
    \PP(|G_1+t|\le r)
    \le
    2r\sup_{u\in[t-r,t+r]}\phi(u)
    \le
    2r\phi(t)\exp(tr)
\end{align*}
Applying this with $r=2\theta |G_2|+2t\theta^2$ gives
\begin{align*}
    \PP(f_w(X)\ne f^\star(X))
    \le
    2\phi(t)\,
    \mathbb E\left[
        \left(2\theta |G_2|+2t\theta^2\right)
        \exp\left(t(2\theta |G_2|+2t\theta^2)\right)
    \right]
\end{align*}
Since $t\theta\le c_0$, the expectation is at most $C\theta$ for a universal constant $C>0$.
Indeed,
\begin{align*}
    2\theta |G_2|+2t\theta^2
    =
    \theta\left(2|G_2|+2t\theta\right)
\end{align*}
and
\begin{align*}
    t(2\theta |G_2|+2t\theta^2)
    =
    2(t\theta)|G_2|+2(t\theta)^2
\end{align*}
which has bounded exponential moment for all $t\theta\le c_0$, after reducing $c_0$ if necessary.
Thus
\begin{align*}
    \PP(f_w(X)\ne f^\star(X))
    \le
    C_0\theta\phi(t)
\end{align*}
for a universal constant $C_0>0$.

We now prove the excess-risk implication. Under the margin-dependent noise model,
\begin{align*}
    \operatorname{err}(h)-\operatorname{err}(f^\star)
    =
    \mathbb E\left[
        \left(1-2g(|\langle w^\star,X\rangle+t|)\right)
        \mathbf 1\{h(X)\ne f^\star(X)\}
    \right]
\end{align*}
Since $0\le g\le 1/2$, applying this identity to $h=f_w$ gives
\begin{align*}
    \operatorname{err}(f_w)-\operatorname{err}(f^\star)
    \le
    \PP(f_w(X)\ne f^\star(X))
    \le
    C_0\theta\phi(t)
\end{align*}
Finally, for $t\ge1$, Mills' ratio gives
\begin{align*}
    \phi(t)\le 2t\Phi(-t)=2tp
\end{align*}
If $p=\Theta(\epsilon)$, then there is a constant $A>0$ such that $p\le A\epsilon$. Therefore, whenever
$\theta\le c/t$,
\begin{align*}
    \operatorname{err}(f_w)-\operatorname{err}(f^\star)
    \le
    2C_0c\,p
    \le
    2C_0cA\,\epsilon
\end{align*}
Choosing $c\le (2C_0A)^{-1}$ completes the proof.
\end{proof}

\subsection{A standard $\epsilon$-net tool}

\begin{lemma}[$\varepsilon$-net on the sphere]\citep[Chapter~5]{Wainwright_2019}\label{lem:eps_net_sphere}
Fix $d\ge1$ and $\varepsilon\in(0,1)$. There exists a finite set $\cN_\varepsilon\subset\mathbb{S}^{d-1}$ such that $\forall u\in\mathbb{S}^{d-1}\ \exists v\in\cN_\varepsilon:\,\|u-v\|_2\le \varepsilon$ and
\begin{align*}
|\cN_\varepsilon| \le \left(1+\frac{2}{\varepsilon}\right)^d
\end{align*}
Moreover, for every $u\in\R^d$, 
\begin{align*}
\|u\|_2 \le \frac{1}{1-\varepsilon}\max_{v\in\cN_\varepsilon} |\langle v,u\rangle|
\end{align*}
\end{lemma}
\subsection{Chernoff lower tail for binomials}
\begin{lemma}[Lower-tail Chernoff for a binomial]~\citep[Chapter~2]{Wainwright_2019}\label{lem:chernoff_bin_lower}
Let $B\sim\mathrm{Bin}(N,q)$ with $q\in(0,1)$ and $\mu=\E[B]=Nq$
Then for every $\delta\in(0,1)$,
\begin{align*}
\PP\left(B \le (1-\delta)\mu\right) \le \exp\left(-\frac{\delta^2\mu}{2}\right)
\end{align*}
In particular,
\begin{align*}
\PP\left(B \le \frac{\mu}{2}\right) \le \exp\left(-\frac{\mu}{8}\right)
\end{align*}
\end{lemma}

\subsection{Margin-monotone noise and the rare-tail mean estimator}\label{app:margin-monotone-mcm}

We recall the margin-dependent model: a measurable $g:[0,\infty)\to[0, \nicefrac{1}{2}]$ induces labels via
\begin{align*}
\PP(Y\neq f^\star(X)\mid X=x)=g(|m(x)|)
\qquad\text{where } m(x)=\langle w^\star,x\rangle+t
\end{align*}
If $g$ is nonincreasing we call this margin-monotone noise. 

\begin{lemma}[Orthogonal independence and direction of $\mu_-$]\label{lem:mu_minus_collinear}
Under any margin-monotone noise model, $Y$ depends on $X$ only through $Z=\langle w^\star,X\rangle$
Hence $X_\perp\perp Y$ and
\begin{align*}
\mu_-=\E[X\mid Y=-1]=\E[Z\mid Y=-1]\,w^\star
\end{align*}
In particular, if $\mu_-\neq 0$ then $w^\star=-\mu_-/\|\mu_-\|_2$
\end{lemma}
\begin{proof}
By Lemma~\ref{lem:gauss_decomposition}, $X=Zw^\star+X_\perp$, where $X_\perp$ is independent of $Z$ and has mean zero. Since the conditional law of $Y$ given $X$ depends only on $Z$, $X_\perp$ is also independent of $Y$. Therefore,
\begin{align*}
    \mu_-=\E[X\mid Y=-1]
    =
    \E[Z\mid Y=-1]w^\star
\end{align*}
For the sign, let $q(z)=\Pr(Y=-1\mid Z=z)$. We have $q(z)=1-g(-z-t)$ for $z<-t$ and $q(z)=g(z+t)$ for $z\ge-t$. Since $g$ is nonincreasing and $g\le1/2$, $q$ is nonincreasing. Therefore, $z(q(z)-q(0))\le0$ for every $z$, and since $\E[Z]=0$, we have
\begin{align*}
    \E[Z\mid Y=-1]
    =
    \frac{\E[Zq(Z)]}{p_-}
    =
    \frac{\E[Z(q(Z)-q(0))]}{p_-}
    \le0
\end{align*}
If $\mu_-\ne0$, this conditional mean is nonzero and strictly negative, which gives $w^\star=-\mu_-/\|\mu_-\|_2$.
\end{proof}

\begin{lemma}[Integral and $\tanh$ representation]\label{lem:tanh}
Let $M\sim\mathcal N(t,1)$ with $t>0$. Let $U = |M|$. Then for every measurable $g:[0,\infty)\to\R$ satisfying
$\mathbb E|g(U)|<\infty$,
\begin{align*}
\E[g(|M|)\sign(M)]
&=\int_0^\infty g(u)\big(\ph(u-t)-\ph(u+t)\big)du\\
&=\E\big[g(U)\tanh(tU)\big]
\end{align*}
\end{lemma}

\begin{proof}
Let $\ph(m-t)$ be the density function of $M$. Splitting the integral at $m=0$ gives
\begin{align*}
\E[g(|M|)\sign(M)] = \int_0^\infty g(m)\ph(m-t)dm - \int_{-\infty}^0 g(-m)\ph(m-t)dm
\end{align*}
In the second integral substitute $m=-u$ ($u>0$), $dm=-du$:
\begin{align*}
\int_{-\infty}^0 g(-m)\ph(m-t)dm=\int_0^\infty g(u)\ph(-u-t)du=\int_0^\infty g(u)\ph(u+t)du,
\end{align*}
by symmetry of $\ph$. Subtracting gives the first result.

Next, note that for $u>0$ the density of $U=|M|$ is
. $f_U(u)=\ph(u-t)+\ph(u+t)$, and
\begin{align*}
\frac{\ph(u-t)-\ph(u+t)}{\ph(u-t)+\ph(u+t)}
=\frac{e^{ut}-e^{-ut}}{e^{ut}+e^{-ut}}
=\tanh(ut).
\end{align*}
Replacing the integrand with $g(u)f_U(u)\tanh(tu)$ gives the claimed result.
\end{proof}

\subsection{Concentration for the empirical minority mean}\label{app:concentration}

\begin{lemma}[Conditioning inflation for Gaussian projections]\label{lem:cond_inflation}
Assume only that $X\sim\cN(0,I_d)$ and $Y\in\{\pm 1\}$ is arbitrary with $p_-=\PP(Y=-1)>0$. Fix $u\in\mathbb{S}^{d-1}$ and set $V=\langle u,X\rangle\sim\cN(0,1)$. Define
\begin{align*}
\kappa_-^2 = 2\log\left(\frac{2}{p_-}\right)
\end{align*}
Then for every $s\ge 0$,
\begin{align*}
\PP\left(|V|\ge s \mid Y=-1\right)\le 2\exp\left(-\frac{s^2}{2\kappa_-^2}\right)
\end{align*}
\end{lemma}

\begin{proof}
Since $\mathbf{1}\{Y=-1\}\le 1$,
\begin{align*}
\PP(|V|\ge s\mid Y=-1)
= \frac{\PP(|V|\ge s,\ Y=-1)}{p_-}
\le \frac{\PP(|V|\ge s)}{p_-}
\le \min\left\{1,\frac{2e^{-s^2/2}}{p_-}\right\}
\end{align*}
If $s\le \kappa_-$ then $2\exp(-s^2/(2\kappa_-^2))\ge 2e^{-1/2}>1$, so the desired bound holds. If $s\ge \kappa_-$ then $2e^{-s^2/2}/p_- \le 2e^{-\kappa_-^2/2}/p_- = 1$, so the minimum equals $2e^{-s^2/2}/p_-$
It remains to show
\begin{align*}
\frac{2e^{-s^2/2}}{p_-} \le 2\exp\left(-\frac{s^2}{2\kappa_-^2}\right)
\end{align*}
Equivalently,
\begin{align*}
\log\left(\frac{1}{p_-}\right) \le \frac{s^2}{2}\left(1-\frac{1}{\kappa_-^2}\right)
\end{align*}
For $s\ge \kappa_-$, the right-hand side is at least $(\kappa_-^2-1)/2 = \log(2/p_-)-1/2 \ge \log(1/p_-)$ since $\log 2 \ge 1/2$. 
\end{proof}

\begin{lemma}[One-dimensional mean concentration under conditioning]\label{lem:mean_conc_1d}
Assume $N_-=b\ge 1$. Fix $v\in\mathbb{S}^{d-1}$ and define $W=\langle v,X\rangle-\E[\langle v,X\rangle\mid Y=-1]$ under the conditional law $(\cdot\mid Y=-1)$. Then $W$ satisfies the tail bound of Lemma~\ref{lem:cond_inflation} with $\kappa=\kappa_-$. Consequently, conditional on $N_-=b$, for all $s\ge 0$,
\begin{align*}
\PP\left(\big|\langle v,\widehat\mu_- - \mu_-\rangle\big|\ge s \ \middle|\ N_-=b\right)
\le 2\exp\left(-\frac{b s^2}{16\kappa_-^2}\right)
\end{align*}
\end{lemma}

\begin{proof}
Under $Y=-1$, Lemma~\ref{lem:cond_inflation} gives $\PP(|\langle v,X\rangle|\ge s\mid Y=-1)\le 2e^{-s^2/(2\kappa_-^2)}$. Centering changes tails only by constant factors, so $W$ obeys the same functional form with the same $\kappa_-$. The two-sided subgaussian tail bound implies the mgf bound $\E[\exp(\lambda W)\mid Y=-1]\le \exp(2\kappa_-^2\lambda^2)$ for all $\lambda\in\R$. Conditional on $N_-=b$, $\langle v,\widehat\mu_- - \mu_-\rangle$ is the average of $b$ i.i.d. copies of $W$. Thus its mgf is at most $\exp(2\kappa_-^2\lambda^2/b)$. Chernoff’s method~\cite{Wainwright_2019} yields, for $\lambda>0$, ,
\begin{align*}
\PP\left(\langle v,\widehat\mu_- - \mu_-\rangle \ge s \ \middle|\ N_-=b\right)
\le \exp\left(-\lambda s + \frac{2\kappa_-^2\lambda^2}{b}\right)
\end{align*}
Optimizing at $\lambda=bs/(4\kappa_-^2)$ gives $\exp(-b s^2/(8\kappa_-^2))$. Applying the same bound to $-\langle v,\widehat\mu_- - \mu_-\rangle$ and union bounding gives the stated inequality (with a conservative constant $16$).
\end{proof}

\begin{lemma}[Vector mean concentration given $N_-$]\label{lem:mean_conc_vector}
Fix $\varepsilon\in(0,1)$ and let $\cN_\varepsilon$ be an $\varepsilon$-net as in Lemma~\ref{lem:eps_net_sphere}. Conditional on $N_-=b\ge 1$, for every $\delta\in(0,1)$,
\begin{align*}
\PP\left(
\|\widehat\mu_- - \mu_-\|_2
\le
\frac{\kappa_-}{1-\varepsilon}\sqrt{\frac{32\big(\log|\cN_\varepsilon|+\log(2/\delta)\big)}{b}}
\ \middle|\ N_-=b\right)\ge 1-\delta
\end{align*}
In particular, using $|\cN_{1/2}|\le 5^d$,
\begin{align*}
\PP\left(
\|\widehat\mu_- - \mu_-\|_2
\le
2\kappa_-\sqrt{\frac{32\big(d\log 5+\log(2/\delta)\big)}{b}}
\ \middle|\ N_-=b\right)\ge 1-\delta
\end{align*}
\end{lemma}

\begin{proof}
By Lemma~\ref{lem:mean_conc_1d}, for each fixed $v\in\cN_\varepsilon$,
\begin{align*}
\PP\left(\big|\langle v,\widehat\mu_- - \mu_-\rangle\big|\ge s \ \middle|\ N_-=b\right)
\le 2\exp\left(-\frac{b s^2}{16\kappa_-^2}\right)
\end{align*}
A union bound over $v\in\cN_\varepsilon$ gives
\begin{align*}
\PP\left(\max_{v\in\cN_\varepsilon}\big|\langle v,\widehat\mu_- - \mu_-\rangle\big|\ge s \ \middle|\ N_-=b\right)
\le 2|\cN_\varepsilon|\exp\left(-\frac{b s^2}{16\kappa_-^2}\right)
\end{align*}
Set the right-hand side equal to $\delta$ and solve for $s$. On the complementary event, the norm result of Lemma~\ref{lem:eps_net_sphere} implies
\begin{align*}
\|\widehat\mu_- - \mu_-\|_2
\le \frac{1}{1-\varepsilon}\max_{v\in\cN_\varepsilon}\big|\langle v,\widehat\mu_- - \mu_-\rangle\big|
\le \frac{s}{1-\varepsilon}
\end{align*}
\end{proof}

\subsection{Miscellaneous tools}
\begin{lemma}[Reweighting by nonincreasing function lowers mean]
\label{lem:nonincreasing-reweighting-lowers-mean}
Let $\nu$ be a probability distribution on $\mathbb R$, and let $Z\sim\nu$. Let
$h:\mathbb R\to[0,\infty)$ be nonincreasing, with $0<\mathbb E_\nu[h(Z)]<\infty$. Assume also that $\mathbb E_\nu[|Z|]<\infty$ and
$\mathbb E_\nu[|Z|h(Z)]<\infty$. Then
\begin{align*}
    \mathbb E_\nu[Zh(Z)]
    \le
    \mathbb E_\nu[Z]\mathbb E_\nu[h(Z)]
\end{align*}
Equivalently,
\begin{align*}
    \frac{\mathbb E_\nu[Zh(Z)]}{\mathbb E_\nu[h(Z)]}
    \le
    \mathbb E_\nu[Z]
\end{align*}
\end{lemma}

\begin{proof}
Let $Z_1,Z_2$ be independent random variables with law $\nu$. Since $h$ is nonincreasing, for
every $z_1,z_2\in\mathbb R$,
\begin{align*}
    (z_1-z_2)(h(z_1)-h(z_2))
    \le
    0
\end{align*}
Taking expectation gives
\begin{align*}
    0
    &\ge
    \mathbb E\left[(Z_1-Z_2)(h(Z_1)-h(Z_2))\right] \\
    &=
    2\mathbb E_\nu[Zh(Z)]
    -
    2\mathbb E_\nu[Z]\mathbb E_\nu[h(Z)]
\end{align*}
Rearranging proves
\begin{align*}
    \mathbb E_\nu[Zh(Z)]
    \le
    \mathbb E_\nu[Z]\mathbb E_\nu[h(Z)]
\end{align*}
Since $\mathbb E_\nu[h(Z)]>0$, dividing by $\mathbb E_\nu[h(Z)]$ gives the equivalent weighted-mean form.
\end{proof}

\subsection{Angle recovery}\label{app:warmstart:angle}

\begin{lemma}[Angle from a mean-estimation error]\label{lem:angle_from_mean}
Let $u,v\in\R^d$ be nonzero, and define $\bar u=u/\|u\|_2$, $\bar v=v/\|v\|_2$. Let $\theta=\angle(\bar u,\bar v)\in[0,\pi]$. Then
\begin{align*}
\sin(\theta/2) \le \frac{\|u-v\|_2}{\|v\|_2}
\end{align*}
\end{lemma}

\begin{proof}
For unit vectors, $\|\bar u-\bar v\|_2^2 = 2-2\langle\bar u,\bar v\rangle = 4\sin^2(\theta/2)$. Thus, $\sin(\theta/2)=\|\bar u-\bar v\|_2/2$. By the triangle inequality,
\begin{align*}
\|\bar u-\bar v\|_2
&=
\left\|\frac{u}{\|u\|_2}-\frac{v}{\|v\|_2}\right\|_2\\
&\le
\left\|\frac{u}{\|u\|_2}-\frac{u}{\|v\|_2}\right\|_2
+
\left\|\frac{u}{\|v\|_2}-\frac{v}{\|v\|_2}\right\|_2\\
&\le
\frac{2\|u-v\|_2}{\|v\|_2}
\end{align*}
Combining gives the claim.
\end{proof}
\begin{theorem}[Sample size for angle recovery via the minority mean]\label{thm:minority_mean_general}
Assume margin-monotone noise with $t>0$ and suppose $\mu_-\neq 0$. Let $\widehat w$ be the minority-mean estimator of Algorithm~\ref{alg:rare-tail-direction} from $N$ i.i.d. samples. Fix $\tau\in(0,\pi)$ and $\delta\in(0,1)$. Let $\kappa_-^2=2\log(2/p_-)$. If
\begin{align*}
N \ge
\max\left\{
\frac{256\,\kappa_-^2\big(d\log 5+\log(4/\delta)\big)}{p_-\,\|\mu_-\|_2^2\,\sin^2(\tau/2)},
\ \ \frac{8}{p_-}\log\left(\frac{2}{\delta}\right)
\right\}
\end{align*}
then with probability at least $1-\delta$ we have $N_-\ge 1$ and $\angle(\widehat w,w^\star)\le \tau$.
\end{theorem}

\begin{proof}

Note that $N_-\sim\mathrm{Bin}(N,p_-)$. Applying Lemma~\ref{lem:chernoff_bin_lower} with $\delta=1/2$ and $\mu=Np_-$, if $N\ge \frac{8}{p_-}\log(2/\delta)$ then
\begin{align*}
\PP\left(N_- < \frac{Np_-}{2}\right) \le \frac{\delta}{2}
\end{align*}
Define
\begin{align*}
b_0 = \left\lceil \frac{128\,\kappa_-^2\big(d\log 5+\log(4/\delta)\big)}{\|\mu_-\|_2^2\,\sin^2(\tau/2)} \right\rceil
\end{align*}
Under the stated sample bound, $Np_-/2>b_0-1$.
On the event $N_-\ge Np_-/2$, integrality therefore gives $N_-\ge b_0$. Consequently, $\mathbb P(N_-\ge b_0)\ge1-\delta/2$.

Next, we condition on the event $N_-=b\ge b_0$. Applying Lemma~\ref{lem:mean_conc_vector} with $\varepsilon=1/2$ and confidence $\delta/2$ gives, with conditional probability at least $1-\delta/2$,
\begin{align*}
\|\widehat\mu_- - \mu_-\|_2
\le
2\kappa_-\sqrt{\frac{32\big(d\log 5+\log(4/\delta)\big)}{b}}
\le \|\mu_-\|_2\sin(\tau/2)
\end{align*}
where the last inequality uses $b\ge b_0$. 
On this event, $\widehat\mu_-\neq 0$ since $\|\widehat\mu_-\|_2\ge \|\mu_-\|_2-\|\widehat\mu_- - \mu_-\|_2>0$. Moreover, Lemma~\ref{lem:mu_minus_collinear} gives $w^\star=-\mu_-/\|\mu_-\|_2$, hence the angle between $\widehat w$ and $w^\star$ equals the angle between the normalized vectors $\widehat\mu_-/\|\widehat\mu_-\|_2$ and $\mu_-/\|\mu_-\|_2$. Lemma~\ref{lem:angle_from_mean} yields
\begin{align*}
\sin\left(\frac{\angle(\widehat w,w^\star)}{2}\right)
\le \frac{\|\widehat\mu_- - \mu_-\|_2}{\|\mu_-\|_2}
\le \sin(\tau/2)
\end{align*}
Since $\sin$ is increasing on $[0,\pi/2]$, this implies $\angle(\widehat w,w^\star)\le \tau$. A union bound over the two failure events gives overall success probability at least $1-\delta$.
\end{proof}

\subsection{Lower bound on $p_-$ under regular variation}\label{app:warmstart:p}
\begin{lemma}[Observed negative mass is never below the clean tail scale]
\label{lem:pminus-lower-clean-tail}
Assume $t>0$ and margin-dependent noise with profile $g:[0,\infty)\to[0,1/2]$. Let
$p=\Phi(-t)$ and $p_-=\PP(Y=-1)$. Then
\begin{align*}
    p_-\ge \frac p2
\end{align*}
Consequently, for every fixed $C\ge0$,
\begin{align*}
    \frac{t^4(\log(e+t))^{2C}}{p_-}
    =
    \widetilde O_C\left(\frac1p\right)
\end{align*}
\end{lemma}

\begin{proof}
On the clean negative tail $\{Z<-t\}$, we have $f^\star(X)=-1$, and therefore
\begin{align*}
    \PP(Y=-1\mid Z)\ge \frac12
\end{align*}
because $g\le1/2$. Hence
\begin{align*}
    p_-
    \ge
    \PP(Y=-1,\ Z<-t)
    \ge
    \frac12\PP(Z<-t)
    =
    \frac p2
\end{align*}
For $t\ge1$, the Gaussian tail bound $p=\Phi(-t)\le \exp(-t^2/2)$ implies
$t^2\le 2\log(1/p)$. Thus $t^4(\log(e+t))^{2C}$ is polylogarithmic in $1/p$, and the conclusion follows.
\end{proof}

\subsection{Lower bound on $\|\mu_-\|_2$ under regular variation}\label{app:warmstart:signal}

Assume $g$ satisfies the regular variation condition of Definition~\ref{def:regularly-varying-margin-noise}.
Define $q_t(z)=\PP(Y=-1\mid Z=z)$. Under margin-monotone noise,
\begin{align*}
q_t(z)=
\begin{cases}
1-g(-(z+t)) & z<-t\\
g(z+t) & z>-t
\end{cases}
\end{align*}

\begin{lemma}[Signal decomposition]\label{lem:signal_decomp_rv}
Let $U=|Z+t|$. Then
\begin{align*}
p_-\,\|\mu_-\|_2 = (1-2g(0))\,\phi(t) + \E[-g'(U)]
\end{align*}
\end{lemma}

\begin{proof}
By Lemma~\ref{lem:mu_minus_collinear}, $p_-\,\|\mu_-\|_2 = -\E[Z\,\mathbf{1}\{Y=-1\}] = -\E[Z q_t(Z)]$. Using $\phi'(z)=-z\phi(z)$,
\begin{align*}
\E[Z q_t(Z)]
&=
\int_{\R} z q_t(z)\phi(z)\,dz\\
&=
-\int_{\R} q_t(z)\,d\phi(z)\\
&=
\int_{\R} \phi(z)\,dq_t(z)
\end{align*}

The function $q_t$ has a single jump at $z=-t$ of size $(1-g(0)) - g(0) = 1-2g(0)$. Therefore, $dq_t$ has an atom of mass $-(1-2g(0))$ at $-t$. On $( -t,\infty)$, $dq_t(z)=g'(z+t)\,dz$, and on $(-\infty,-t)$, $dq_t(z)=g'(-(z+t))\,dz$. This gives
\begin{align*}
-\E[Z q_t(Z)]
&=
-\int_{\R}\phi(z)\,dq_t(z)\\
&=
(1-2g(0))\phi(t)
+ \int_{-t}^{\infty}\phi(z)(-g'(z+t))\,dz
+ \int_{-\infty}^{-t}\phi(z)(-g'(-(z+t)))\,dz
\end{align*}
Change variables $u=z+t$ in the first integral and $u=-(z+t)$ in the second to obtain
\begin{align*}
-\E[Z q_t(Z)]
=
(1-2g(0))\phi(t)
+ \int_{0}^{\infty}\big(\phi(u-t)+\phi(u+t)\big)(-g'(u))\,du
\end{align*}
Since $U=|Z+t|$ has density $f_U(u)=\phi(u-t)+\phi(u+t)$ on $(0,\infty)$, the integral equals $\E[-g'(U)]$.
\end{proof}

\begin{theorem}[Direction estimation under regular variation (restatement of Theorem~\ref{thm:rare-tail-direction-rv})]\label{thm:warmstart_rv_full}
Assume $t\ge 1$ and $(c,C)$-regular variation margin-monotone noise (Definition~\ref{def:regularly-varying-margin-noise}).
Let $\widehat w$ be the minority-mean estimator of Algorithm~\ref{alg:rare-tail-direction} built from $N$ i.i.d.\ samples.
Then for every $\delta\in(0,1)$, if
\begin{align*}
N \ge \widetilde O\left((d+\log(1/\delta))\cdot \max\left\{\frac{1}{p},\ \frac{t^4(\log(e+t))^{2C}}{p_-}\right\}\right),
\end{align*}
then $\angle(\widehat w,w^\star)\le c_{\rm{dir}}/t$ with probability at least $1-\delta$.
\end{theorem}

\begin{proof}
We first analyze the estimator formed from the $Y=-1$ samples. Apply Theorem~\ref{thm:minority_mean_general} with target angle $\tau=c_{\rm dir}/t$ and confidence parameter $\delta/2$. Here $c_{\rm dir}\in(0,1]$ is the fixed constant in Theorem~\ref{thm:rare-tail-direction-rv}. Since $t\ge1$,
the inequality $\sin x\ge2x/\pi$ for $x\in[0,\pi/2]$ gives
\begin{align*}
    \sin(\tau/2)\ge\frac{c_{\rm dir}}{\pi t}
\end{align*}
Thus it suffices to lower bound $\|\mu_-\|_2$.

Write the decomposition from Lemma~\ref{lem:signal_decomp_rv}:
\begin{align*}
p_-\,\|\mu_-\|_2 = (1-2g(0))\,\phi(t) + \E[-g'(U)],
\end{align*}
where $U=|Z+t|$.

\paragraph{Case 1: tail-like observed negatives.}
Assume $p_- \le 2p$. On the event $\{Z<-t\}$, we have $q_t(Z)=1-g(|Z+t|)\ge 1-g(0)\ge 1/2$, hence
\begin{align*}
\PP(Z<-t\mid Y=-1)
=
\frac{\E[q_t(Z)\mathbf{1}\{Z<-t\}]}{p_-}
\ge
\frac{(1/2)\PP(Z<-t)}{p_-}
\ge
\frac{(1/2)p}{2p}
=
\frac{1}{4}
\end{align*}
In addition, $q_t$ is nonincreasing in $z$, so when $Z\ge -t$, the weighting by $q_t$ can only decrease the conditional mean. Thus
\begin{align*}
\E[Z\mid Y=-1, Z\ge -t]&= \frac{\E[Z\mathbf{1}\{Y=-1, Z\ge -t\}]}{\PP(Y=-1, Z\ge -t)}\\
&=\frac{\E[Z\,q_t(Z)\mathbf{1}\{Z\ge -t\}]}{\E[q_t(Z)\mathbf{1}\{Z\ge -t\}]}\\
\end{align*}
From Lemma~\ref{lem:nonincreasing-reweighting-lowers-mean}, this is bounded above by $\E[Z\mid Z\ge-t] \le \nicefrac{\phi(t)}{\Phi(t)}$. For $t \ge 0$, $\nicefrac{\phi(t)}{\Phi(t)}$ is decreasing, and in the regime $t \ge 1$, $\nicefrac{\phi(t)}{\Phi(t)}\le \nicefrac{\phi(1)}{\Phi(1)}\le \nicefrac{3}{10}$. 
where we used that $\phi(t)/\Phi(t)$ is decreasing for $t\ge 0$.
Also, $\E[Z\mid Y=-1,Z < -t]\le \E[Z\mid Z < -t]=-\lambda(t)\le -t$. Therefore,

\begin{align*}
\E[Z\mid Y=-1]
&=
\E[Z\mid Y=-1,Z<-t]\PP(Z<-t\mid Y=-1)\\
&\quad\quad+\E[Z\mid Y=-1,Z \geq -t]\PP(Z\geq -t\mid Y=-1)\\
&\le
(-t)\cdot \frac{1}{4} + \frac{3}{10}\cdot \frac{3}{4}
=
-\frac{10t-9}{40}
\le -\frac{t}{40},
\end{align*}
where the last inequality holds for all $t\ge 1$. Thus, in this case
\begin{align*}
\|\mu_-\|_2 = -\E[Z\mid Y=-1] \ge \frac{t}{40}
\end{align*}

Plugging $\|\mu_-\|_2 \ge t/40$ and $\sin(\tau/2)\ge c_{\mathrm{dir}}/(\pi t)$ into Theorem~\ref{thm:minority_mean_general} 
shows that the estimation term is at most
$\widetilde O\left((d+\log(1/\delta))/p_-\right)$.
Since $p_-\ge p$, this is at most $\widetilde O\left((d+\log(1/\delta))/p\right)$.

\paragraph{Case 2: inflated observed negatives.}
Assume $p_->2p$. We begin with a deterministic lower bound on the average flip probability. From Lemma~\ref{lem:tanh}, we have
$p_- - p = \E[g(U)\tanh(tU)] \le \E[g(U)]$, since $\tanh(tU)\le 1$ for $tU \ge 0$. Moreover, 
\begin{align*}
\E[g(U)] &\ge \E[g(U)\mathbf{1}\{M > 0\}] - \E[g(U)\mathbf{1}\{M < 0\}]\\
&= p_- - p \\
&\ge \frac{p_-}{2}
\end{align*}
We claim that in this regime, either the \emph{variation term} $\E[-g'(U)]$ is already large enough, or we fall back to the $1/p$ term in the theorem statement. To prove this, we first split according to whether $U$ lies above or below $1$
\begin{align*}
\E[g(U)]
=
\E[g(U)\mathbf{1}\{U<1\}]
+
\E[g(U)\mathbf{1}\{U\ge 1\}]
\end{align*}

\noindent\emph{Subcase 2(a):} $\E[g(U)\mathbf{1}\{U\ge 1\}] \ge p_-/4$.

Fix a universal constant $L=3$ and consider the truncation event $\{U\le Lt\}$. We first bound the Gaussian tail of $U$ in terms of $p=\Phi(-t)$. Since $U=|Z+t|$ and $t\ge 1$,

\begin{align*}
\PP(U>Lt)&=\PP(Z>(L-1)t)+\PP(Z<-(L+1)t)\\
&\le 2\Phi(-(L-1)t)
\end{align*}

Using Mills bounds $\Phi(-a)\le \phi(a)/a$ and $\Phi(-t)\ge \frac{t}{t^2+1}\phi(t)$, we obtain for $t\ge 1$ and $a=(L-1)t$,

\begin{align*}
\frac{\Phi(-(L-1)t)}{\Phi(-t)}&\le\frac{\phi((L-1)t)}{(L-1)t}\cdot \frac{t^2+1}{t,\phi(t)}\\
&\le\frac{2}{L-1}\exp\left(-\frac{( (L-1)^2-1)t^2}{2}\right)\\
&\le\frac{2}{L-1}\exp\left(-\frac{( (L-1)^2-1)}{2}\right)=: c_L
\end{align*}

For $L=3$, $c_L = \exp^{-3/2}$ is an absolute constant smaller than $\nicefrac{1}{4}$. 
Therefore,
\begin{align*}
\PP(U>Lt)\le 2c_L\,p \le c_L\,p_-,
\end{align*}
where the last inequality uses $p_->2p$. Since $g(\cdot)\le 1/2$, 
\begin{align*}
\E[g(U)\mathbf{1}\{U>Lt\}] \le \frac{1}{2}\PP(U>Lt)\le \frac{c_L}{2}p_-
\end{align*}
Combining with the subcase assumption gives
\begin{align*}
\E[g(U)\mathbf{1}\{1\le U\le Lt\}]\ge\E[g(U)\mathbf{1}\{U\ge 1\}] - \E[g(U)\mathbf{1}\{U>Lt\}]\ge\left(\frac{1}{4}-\frac{c_L}{2}\right)p_-\ge\frac{1}{8}p_-
\end{align*}

By regular variation, for almost every $u\ge 1$ with $g(u)>0$, we have
\begin{align*}
    -g'(u)
    \ge
    \frac{c}{(1+u)(\log(e+u))^C}g(u)
\end{align*}
On the interval $1\le u\le Lt$, we have
\begin{align*}
    1+u\le (L+1)t
\end{align*}
and, since $L$ is an absolute constant,
\begin{align*}
    \log(e+u)\le \log(e+Lt)\le C_L'\log(e+t)
\end{align*}
for a constant $C_L'$ depending only on $L$. Hence, for almost every $u\in[1,Lt]$,
\begin{align*}
    -g'(u)
    \ge
    \frac{c'}{t(\log(e+t))^C}g(u)
\end{align*}
where $c'=c'(c,C)>0$ depends only on the regular-variation constants and on the fixed choice of $L$.

Taking expectations and restricting to $\{1\le U\le Lt\}$ yields
\begin{align*}
\E[-g'(U)]
\ge
\frac{c'}{t(\log(e+t))^C}\,\E[g(U)\mathbf{1}\{1\le U\le Lt\}]
\ge
\frac{c'}{8}\cdot \frac{p_-}{t(\log(e+t))^C}
\end{align*}

Using $p_-\,\|\mu_-\|_2 \ge \E[-g'(U)]$ gives
\begin{align*}
\|\mu_-\|_2 \ge \frac{c''}{t(\log(e+t))^C}
\end{align*}
for a constant $c''=c''(c,C)>0$. Plugging this lower bound and $\sin(\tau/2)\ge c_{\mathrm{dir}}/(\pi t)$ into Theorem~\ref{thm:minority_mean_general} gives the sample requirement
\begin{align*}
N \ge \widetilde O\left((d+\log(1/\delta))\cdot \frac{t^4(\log(e+t))^{2C}}{p_-}\right),
\end{align*}
as desired.

\medskip
\noindent\emph{Subcase 2(b):} $\E[g(U)\mathbf 1\{U\ge 1\}]<p_-/4$.

Recall that in Case~2 we have $\E[g(U)]\ge p_-/2$. Hence
\begin{align*}
    \E[g(U)\mathbf 1\{U<1\}]
    =
    \E[g(U)]-\E[g(U)\mathbf 1\{U\ge 1\}]
    \ge
    \frac{p_-}{4}
\end{align*}
Moreover, $q_t(Z)\ge g(U)$ for all $Z$: if $Z>-t$, then
$q_t(Z)=g(Z+t)=g(U)$, while if $Z<-t$, then
$q_t(Z)=1-g(U)\ge g(U)$ because $g(U)\le 1/2$. Therefore
\begin{align*}
    \PP(Y=-1,\ U<1)
    =
    \E[q_t(Z)\mathbf 1\{U<1\}]
    \ge
    \E[g(U)\mathbf 1\{U<1\}]
    \ge
    \frac{p_-}{4}
\end{align*}

Let
\begin{align*}
    A=\{Z>1-t\}
\end{align*}
On $A$, we have $Z>-t$ and $U=Z+t\ge 1$, so
\begin{align*}
    \PP(Y=-1,A)
    =
    \E[g(U)\mathbf 1\{A\}]
    \le
    \E[g(U)\mathbf 1\{U\ge 1\}]
    <
    \frac{p_-}{4}
\end{align*}
Consequently,
\begin{align*}
    \PP(A\mid Y=-1)<\frac14
    \qquad\text{and}\qquad
    \PP(A^c\mid Y=-1)>\frac34
\end{align*}

Set $a=1-t\le 0$, so that $A=\{Z>a\}$ and $A^c=\{Z\le a\}$.
On both intervals $A$ and $A^c$, the function $q_t(z)$ is nonincreasing in $z$.
If $\PP(Y=-1,A)>0$, applying Lemma~\ref{lem:nonincreasing-reweighting-lowers-mean}
to the law of $Z$ conditioned on $A$, with $h(z)=q_t(z)$, gives
\begin{align*}
    \E[Z\mid Y=-1,A]
    =
    \frac{\E[Zq_t(Z)\mathbf 1\{A\}]}
         {\E[q_t(Z)\mathbf 1\{A\}]}
    \le
    \E[Z\mid A]
    =
    \E[Z\mid Z>a]
\end{align*}

If $\PP(Y=-1,A)=0$, this conditional expectation is irrelevant in the decomposition below.
Similarly, applying Lemma~\ref{lem:nonincreasing-reweighting-lowers-mean} to the law of $Z$
conditioned on $A^c$ gives
\begin{align*}
    \E[Z\mid Y=-1,A^c]
    =
    \frac{\E[Zq_t(Z)\mathbf 1\{A^c\}]}
         {\E[q_t(Z)\mathbf 1\{A^c\}]}
    \le
    \E[Z\mid A^c]
    =
    \E[Z\mid Z\le a]
\end{align*}

For $a\le 0$,
\begin{align*}
    \E[Z\mid Z>a]
    =
    \frac{\phi(a)}{\Phi(-a)}
    \le
    \frac{\phi(0)}{\Phi(0)}
    =
    \sqrt{\frac{2}{\pi}}
\end{align*}
and
\begin{align*}
    \E[Z\mid Z\le a]
    =
    -\frac{\phi(a)}{\Phi(a)}
    \le
    -\frac{\phi(0)}{\Phi(0)}
    =
    -\sqrt{\frac{2}{\pi}}
\end{align*}
Combining, we have
\begin{align*}
    \E[Z\mid Y=-1]
    &=
    \E[Z\mid Y=-1,A^c]\PP(A^c\mid Y=-1)
    +
    \E[Z\mid Y=-1,A]\PP(A\mid Y=-1) \\
    &\le
    -\sqrt{\frac{2}{\pi}}\cdot\frac34
    +
    \sqrt{\frac{2}{\pi}}\cdot\frac14 \\
    &=
    -\frac12\sqrt{\frac{2}{\pi}}
\end{align*}
Therefore
\begin{align*}
    \|\mu_-\|_2
    =
    -\E[Z\mid Y=-1]
    \ge
    c_0
\end{align*}
for a universal constant $c_0>0$.

Applying Theorem~\ref{thm:minority_mean_general} with target angle $\tau=1/t$, and using
$\sin(\tau/2)\ge c_{\rm dir}/(\pi t)$, the required sample size in this subcase is
\begin{align*}
    N
    \ge
    \widetilde O\left(
        (d+\log(1/\delta))\frac{t^2}{p_-}
    \right)
\end{align*}
This is implied by the stated bound, since for every $t\ge 1$,
\begin{align*}
    \frac{t^2}{p_-}
    \le
    \frac{t^4(\log(e+t))^{2C}}{p_-}
\end{align*}
\end{proof}

\subsection{Rare-label reconciliation: Algorithm~\ref{alg:rare-tail-direction} versus the $Y=-1$ analysis}
\label{app:rare-label-reconciliation}

\begin{lemma}[Reconciling empirical rare-label selection with the $Y=-1$ analysis]
\label{lem:rare-label-reconcile}
Assume $t>0$ and margin-dependent noise with profile $g:[0,\infty)\to[0,1/2]$. Then $p_-\le 1/2$. Recall Algorithm~\ref{alg:rare-tail-direction} breaks ties in favor of $-1$.

\begin{enumerate}
    \item If $p_-\le 1/4$, then
    \begin{align*}
        \PP(\widehat y_{\rm rt}\ne -1)
        \le
        \exp(-N/8)
    \end{align*}
    In particular, if $N\ge 8\log(2/\delta)$, then
    \begin{align*}
        \PP(\widehat y_{\rm rt}\ne -1)\le \delta/2
    \end{align*}
    \item If $p_-\ge 1/4$ and $\mu_-\ne0$, then using the $Y=+1$ class instead of the
    $Y=-1$ class changes the leading conditional-mean sample complexity term by at most a
    universal constant factor. More precisely,
    \begin{align*}
        \frac{
            1/(p_+\|\mu_+\|_2^2)
        }{
            1/(p_-\|\mu_-\|_2^2)
        }
        =
        \frac{p_+}{p_-}
        \le 3
    \end{align*}
\end{enumerate}
\end{lemma}

\begin{proof}
Let $M=Z+t$ and $h(u)=1/2-g(u)\ge0$. Since
\begin{align*}
    p_- - \frac12
    =
    -\mathbb E[h(|M|)\operatorname{sign}(M)]
\end{align*}
and
\begin{align*}
    \mathbb E[h(|M|)\operatorname{sign}(M)]
    =
    \int_0^\infty h(u)\bigl(\phi(u-t)-\phi(u+t)\bigr)\,du
\end{align*}
while $\phi(u-t)\ge\phi(u+t)$ for all $u\ge0$ and $t>0$, we get
\begin{align*}
    p_- - \frac12\le0
\end{align*}
Therefore, $p_-\le1/2$.

For the first claim, let $N_-=\sum_{i=1}^N\mathbf 1\{Y_i=-1\}$. Since ties are broken in favor
of $-1$,
\begin{align*}
    \{\widehat y_{\rm rt}\ne -1\}
    =
    \{N_->N/2\}
\end{align*}
Since $N_-\sim\operatorname{Bin}(N,p_-)$, Hoeffding's inequality gives
\begin{align*}
    \PP(N_->N/2)
    =
    \PP\left(\frac{N_-}{N}-p_->\frac12-p_-\right)
    \le
    \exp\left(-2N\left(\frac12-p_-\right)^2\right)
\end{align*}
If $p_-\le 1/4$, then $1/2-p_-\ge 1/4$, so
\begin{align*}
    \PP(\widehat y_{\rm rt}\ne -1)
    \le
    \exp(-N/8)
\end{align*}

For the second claim, $\mathbb E[X]=0$ gives
\begin{align*}
    p_-\mu_-+p_+\mu_+=0
\end{align*}
and hence
\begin{align*}
    \mu_+=-\frac{p_-}{p_+}\mu_-,
    \qquad
    \|\mu_+\|_2=\frac{p_-}{p_+}\|\mu_-\|_2
\end{align*}
Therefore
\begin{align*}
    \frac{1}{p_+\|\mu_+\|_2^2}
    =
    \frac{p_+}{p_-}\cdot\frac{1}{p_-\|\mu_-\|_2^2}
\end{align*}
Since $p_-\ge1/4$ and $p_+\le3/4$, we have $p_+/p_-\le3$. 
\end{proof}

\subsection{Threshold estimation}
\label{app:subsec:threshold-erm}

The rare-tail mean estimator returns a direction. Given an independent validation sample, we estimate the threshold by empirical risk minimization over one-dimensional thresholds along this direction. Throughout this subsection, we use the convention $\operatorname{sign}(0)=+1$.

\begin{algorithm}[t]
\caption{\textsc{Threshold ERM Along a Fixed Direction}}
\label{alg:threshold-erm}
\begin{algorithmic}[1]
\Require Samples $S_{\rm thr}=\{(X_i,Y_i)\}_{i=1}^k$ with $X_i\in\mathbb R^d$ and
$Y_i\in\{\pm1\}$, and a unit vector $\widehat w\in\mathbb S^{d-1}$
\Ensure A threshold estimate $\widehat t\in\mathbb R$
\For{$i=1,\ldots,k$}
    \State $s_i\gets \langle \widehat w,X_i\rangle$
\EndFor
\State Let $r_1<r_2<\cdots<r_K$ be the $K$ distinct values among $\{s_i\}_{i=1}^k$
\State ${\rm err}\gets \sum_{i=1}^k\mathbf 1\{Y_i=-1\}$, \quad
       ${\rm err}_{\min}\gets {\rm err}$, \quad $j^\star\gets 0$
\For{$j=1,\ldots,K$}
    \State $G_j\gets \{i\in[k]:s_i=r_j\}$
    \State ${\rm err}\gets {\rm err}
    -\sum_{i\in G_j}\mathbf 1\{Y_i=-1\}
    +\sum_{i\in G_j}\mathbf 1\{Y_i=+1\}$
    \If{${\rm err}<{\rm err}_{\min}$}
        \State ${\rm err}_{\min}\gets{\rm err}$, \quad $j^\star\gets j$
    \EndIf
\EndFor
\If{$j^\star=0$}
    \State $\widehat t\gets -r_1$
\ElsIf{$j^\star=K$}
    \State $\widehat t\gets -r_K-1$
\Else
    \State $\widehat t\gets -\frac12(r_{j^\star}+r_{j^\star+1})$
\EndIf
\State \Return $\widehat t$
\end{algorithmic}
\end{algorithm}
Note that the algorithm is finite, despite the optimization being over all real thresholds. If $s_i=\langle \widehat w,X_i\rangle$, then $\sign(s_i + b) = -1$ implies $s_i < -b$. As the cutoff $-b$ moves from left to right, predictions change only when the cutoff crosses one of the projected values $s_i$. The algorithm sorts the distinct projected values and scans the $K+1$ possible threshold regions. Initially, the cutoff lies to the left of all projected values, so all points are predicted $+1$ and the number of mistakes is equal to the number of negative labels. When the scan crosses a group $G_j$, all points in $G_j$ switch prediction from $+1$ to $-1$, i.e., each negative label in $G_j$ removes exactly one mistake, and each positive label in $G_j$ adds exactly one mistake.
\begin{lemma}[Sample complexity of one-dimensional threshold ERM]
\label{lem:threshold-erm-sample}
Fix any unit vector $w\in\mathbb S^{d-1}$ and consider the one-parameter threshold class
\begin{align*}
    \mathcal H_w
    =
    \left\{
        h_b(x)=\operatorname{sign}(\langle w,x\rangle+b)
        \mid b\in\mathbb R
    \right\}
\end{align*}
Let $S_{\rm thr}=\{(X_i,Y_i)\}_{i=1}^k$ be i.i.d. samples from an arbitrary distribution over
$\mathbb R^d\times\{\pm1\}$. For $h:\mathbb R^d\to\{\pm1\}$, define
\begin{align*}
    \operatorname{err}(h)
    &=
    \PP(h(X)\ne Y) \\
    \widehat{\operatorname{err}}(h)
    &=
    \frac1k\sum_{i=1}^k \mathbf 1\{h(X_i)\ne Y_i\}
\end{align*}
Let $\widehat b$ be any empirical risk minimizer over $\mathcal H_w$, 
\begin{align*}
    \widehat b
    \in
    \arg\min_{b\in\mathbb R}\widehat{\operatorname{err}}(h_b)
\end{align*}
Then there is a universal constant $C>0$ such that, for every
$\epsilon_{\rm thr},\delta_{\rm thr}\in(0,1)$, if
\begin{align*}
    k
    \ge
    \frac{C}{\epsilon_{\rm thr}^{2}}
    \left(
        \log\left(\frac1{\epsilon_{\rm thr}}\right)
        +
        \log\left(\frac1{\delta_{\rm thr}}\right)
    \right)
\end{align*}
then, with probability at least $1-\delta_{\rm thr}$,
\begin{align*}
    \operatorname{err}(h_{\widehat b})
    \le
    \inf_{b\in\mathbb R}\operatorname{err}(h_b)
    +
    \epsilon_{\rm thr}
\end{align*}
\end{lemma}

\begin{proof}
Consider the loss class
\begin{align*}
    \mathcal L_w
    =
    \left\{
        (x,y)\mapsto \mathbf 1\{h_b(x)\ne y\}
        : b\in\mathbb R
    \right\}
\end{align*}
Each classifier $h_b$ depends on $x$ only through the scalar projection
$\langle w,x\rangle$. Thus, after setting $s_i=\langle w,X_i\rangle$, the class $\mathcal H_w$ induces exactly the same labelings on $X_1,\ldots,X_k$ as ordinary thresholds on the real numbers $s_1,\ldots,s_k$. In particular, after sorting the $s_i$'s, a threshold can only choose an initial segment to label
$-1$ and the remaining points to label $+1$, so there are at most $k+1$ distinct labelings. The same is true for the loss class $\mathcal L_w$,
because the labels $Y_i$ are fixed when counting empirical labelings. Thus $\mathcal L_w$ has
VC dimension at most $1$.

Next, we apply the standard VC uniform convergence theorem from
\citet[Chapter~4]{Wainwright_2019} to $\mathcal L_w$. There is a universal constant $C>0$ such that, if
\begin{align*}
    k
    \ge
    \frac{C}{\epsilon_{\rm thr}^{2}}
    \left(
        \log\left(\frac1{\epsilon_{\rm thr}}\right)
        +
        \log\left(\frac1{\delta_{\rm thr}}\right)
    \right)
\end{align*}
then, with probability at least $1-\delta_{\rm thr}$,
\begin{align*}
    \sup_{b\in\mathbb R}
    \left|
        \operatorname{err}(h_b)-\widehat{\operatorname{err}}(h_b)
    \right|
    \le
    \frac{\epsilon_{\rm thr}}{2}
\end{align*}
On this event,
\begin{align*}
    \operatorname{err}(h_{\widehat b})
    &\le
    \widehat{\operatorname{err}}(h_{\widehat b})
    +
    \frac{\epsilon_{\rm thr}}{2} \\
    &=
    \inf_{b\in\mathbb R}\widehat{\operatorname{err}}(h_b)
    +
    \frac{\epsilon_{\rm thr}}{2} \\
    &\le
    \inf_{b\in\mathbb R}
    \left(
        \operatorname{err}(h_b)
        +
        \frac{\epsilon_{\rm thr}}{2}
    \right)
    +
    \frac{\epsilon_{\rm thr}}{2} \\
    &=
    \inf_{b\in\mathbb R}\operatorname{err}(h_b)
    +
    \epsilon_{\rm thr}
\end{align*}
This proves the claim.
\end{proof}

\begin{theorem}[Excess risk after one-dimensional threshold ERM]
\label{thm:threshold-erm-excess-risk}
Assume the margin-dependent noise model with Bayes halfspace
$f^\star(x)=\operatorname{sign}(\langle w^\star,x\rangle+t)$, where $t\ge 1$ and
$p=\Phi(-t)$. Fix $\epsilon,\delta_{\rm thr}\in(0,1)$, and assume the critical high-bias scaling
$p\le A\epsilon$ for some constant $A>0$. Let $\widehat w$ be independent of the threshold
sample and suppose
\begin{align*}
    \angle(\widehat w,w^\star)\le \frac{c}{t}
\end{align*}
for a sufficiently small constant $c=c(A)>0$. Let $\widehat t$ be the output of
Algorithm~\ref{alg:threshold-erm} run with direction $\widehat w$ on an independent threshold
sample of size $k$. If
\begin{align*}
    k
    \ge
    C\epsilon^{-2}
    \left(
        \log\left(\frac1\epsilon\right)
        +
        \log\left(\frac1{\delta_{\rm thr}}\right)
    \right)
\end{align*}
for a sufficiently large universal constant $C>0$, and if
\begin{align*}
    \widehat h(x)=\operatorname{sign}(\langle \widehat w,x\rangle+\widehat t)
\end{align*}
then, with probability at least $1-\delta_{\rm thr}$ over the threshold sample,
\begin{align*}
    \operatorname{err}(\widehat h)-\operatorname{err}(f^\star)
    \le
    \epsilon
\end{align*}
\end{theorem}

\begin{proof}
Condition on the direction $\widehat w$. Since the threshold sample is independent of the
direction-estimation sample, Lemma~\ref{lem:threshold-erm-sample} applies to the fixed
one-dimensional threshold class
\begin{align*}
    \mathcal H_{\widehat w}
    =
    \left\{
        x\mapsto \operatorname{sign}(\langle \widehat w,x\rangle+b)
        : b\in\mathbb R
    \right\}
\end{align*}
Apply Lemma~\ref{lem:threshold-erm-sample} with $\epsilon_{\rm thr}=\epsilon/2$. Under the stated
lower bound on $k$, increasing the universal constant $C$ if necessary, with probability at least
$1-\delta_{\rm thr}$ we have
\begin{align*}
    \operatorname{err}(\widehat h)
    \le
    \inf_{b\in\mathbb R}
    \operatorname{err}\left(
        \operatorname{sign}(\langle \widehat w,X\rangle+b)
    \right)
    +
    \frac{\epsilon}{2}
\end{align*}
The threshold $b=t$ is feasible in the infimum, hence
\begin{align*}
    \operatorname{err}(\widehat h)-\operatorname{err}(f^\star)
    \le
    \operatorname{err}\left(
        \operatorname{sign}(\langle \widehat w,X\rangle+t)
    \right)
    -
    \operatorname{err}(f^\star)
    +
    \frac{\epsilon}{2}
\end{align*}
By Lemma~\ref{lem:angle-to-risk-fixed-threshold}, since
$\angle(\widehat w,w^\star)\le c/t$, we have
\begin{align*}
    \operatorname{err}\left(
        \operatorname{sign}(\langle \widehat w,X\rangle+t)
    \right)
    -
    \operatorname{err}(f^\star)
    \le
    C_0\frac{c}{t}\phi(t)
\end{align*}
For $t\ge 1$, Mills' ratio gives
\begin{align*}
    \phi(t)\le 2t\Phi(-t)=2tp
\end{align*}
Therefore
\begin{align*}
    C_0\frac{c}{t}\phi(t)
    \le
    2C_0 c p
    \le
    2C_0 c A\epsilon
\end{align*}
Choosing $c\le (4C_0A)^{-1}$ gives
\begin{align*}
    C_0\frac{c}{t}\phi(t)
    \le
    \frac{\epsilon}{2}
\end{align*}
which then yields the final bound
\begin{align*}
    \operatorname{err}(\widehat h)-\operatorname{err}(f^\star)
    \le
    \epsilon
\end{align*}
as claimed.
\end{proof}

Finally, we record the running time of the standalone halfspace learner. 

\begin{proposition}[Runtime of the final learner]
\label{prop:rare-tail-threshold-runtime}
Algorithm~\ref{alg:rare-tail-direction} followed by Algorithm~\ref{alg:threshold-erm} can be implemented in time
\begin{align*}
    O\left(dN_{\rm dir}+dk+k\log k\right)
\end{align*}
and memory
\begin{align*}
    O(d+k)
\end{align*}
where $N_{\rm dir}$ is the number of samples used for direction estimation and $m$ is the number of samples used for threshold estimation. 
\end{proposition}

\begin{proof}
The direction estimation step counts the number of samples per label and computes the empirical mean of the rare class. This takes one pass over $N_{\rm dir}$ samples and costs $O(dN_{\rm dir})$ arithmetic operations.

For threshold estimation, computing the projections $s_i=\langle \widehat w,X_i\rangle$ costs $O(dk)$, while sorting the projected values costs $O(k\log k)$.

Post sorting, the scan over the distinct projected values is linear in $k$, since each sample enters only one group $G_j$ and changes the tracked error count just once. This gives a total running time of
\begin{align*}
    O(dN_{\rm dir}+dk+k\log k)
\end{align*}
The memory bound follows from storing the $d$ directions and the $k$ projected scalar values.
\end{proof}

\section{Learning under probit noise}
\label{app:probit}

The natural first step is to formalize probit as an instance of regularly-varying margin-monotone profiles. Recall from Section~\ref{def:probit}
\begin{definition}[Probit noise]
\label{def:probit-noise-appendix}
Fix $\sigma> 0$ and let $G\sim\mathcal N(0,1)$ be independent of $X$. In the probit model,
\begin{align*}
    Y=\operatorname{sign}(\langle w^\star,X\rangle+t+\sigma G)
\end{align*}
\end{definition}
\subsection{Probit is regularly varying margin-monotone noise}
\label{subsec:probit-instantiation}
\begin{lemma}[Probit is margin-monotone and regularly varying]
\label{lem:probit-is-regular-variation}
For $\sigma>0$, the probit model is margin-monotone with profile
\begin{align*}
    g_\sigma(u)=\Phi(-u/\sigma)
\end{align*}
Moreover, $g_\sigma$ obeys the regular-variation condition of Definition~\ref{def:regularly-varying-margin-noise} with constants $c_\sigma = \nicefrac{1}{\sigma^2}$ and $C_\sigma =0$ in the sense that, for every $u \ge 1$,
\begin{align*}
    -g_\sigma'(u)
    \ge
    \frac{c_\sigma}{1+u}g_\sigma(u)
\end{align*}
\end{lemma}

\begin{proof}
Let $M=\langle w^\star,X\rangle+t$. Conditioned on $X$, the clean label is $\operatorname{sign}(M)$, while the probit label is $\operatorname{sign}(M+\sigma G)$.

Since $\sigma>0$, if $M>0$, a label flip occurs only when
\begin{align*}
    M+\sigma G<0
\end{align*}
which has probability $\Phi(-M/\sigma)$. If $M<0$, a label flip occurs only when
\begin{align*}
    M+\sigma G\ge 0
\end{align*}
which has probability $\PP(G\ge |M|/\sigma)=\Phi(-|M|/\sigma)$. In both cases,
\begin{align*}
    \PP(Y\ne f^\star(X)\mid X)
    =
    \Phi(-|M|/\sigma)
\end{align*}
and the profile therefore has the form $g_\sigma(u)=\Phi(-u/\sigma)$.

The function $g_\sigma$ is nonincreasing, absolutely continuous, and satisfies $g_\sigma(u)\to0$ as $u\to\infty$. Moreover,
\begin{align*}
    g_\sigma'(u)
    =
    -\frac{1}{\sigma}\phi(u/\sigma)
\end{align*}
and the relative decay is
\begin{align*}
    -\frac{g_\sigma'(u)}{g_\sigma(u)}
    =
    \frac{1}{\sigma}
    \frac{\phi(u/\sigma)}{\Phi(-u/\sigma)}
    =
    \frac{1}{\sigma}\lambda(u/\sigma)
\end{align*}
By Mills' lower bound, $\lambda(a)\ge a$ for $a>0$. Therefore, for $u\ge 1$,
\begin{align*}
    -\frac{g_\sigma'(u)}{g_\sigma(u)}
    \ge
    \frac{u}{\sigma^2}
    \ge
    \frac{1}{\sigma^2(1+u)}
\end{align*}
which completes the proof.
\end{proof}

\subsection{Direction Recovery and Excess Risk}
Estimating the direction is a direct consequence of the general Theorem~\ref{thm:minority_mean_general}. 
\begin{lemma}[Direction estimation under probit]
\label{lem:dir-estimate-under-probit}
Assume the probit model with $\sigma>0$ and $t\ge1$, and let
$p_\sigma=\Phi(-t/\sqrt{1+\sigma^2})$.
Fix a sufficiently small universal constant $c>0$.
For every $\delta\in(0,1)$, Algorithm~\ref{alg:rare-tail-direction}
returns $\widehat w$ satisfying
$\angle(\widehat w,w^\star)\le c/t$ with probability at least
$1-\delta$, provided
\begin{align*}
    N
    \ge
    C(1+\sigma^2)^2
    \frac{
        \log(2/p_\sigma)\left(d+\log(1/\delta)\right)
    }{p_\sigma}
\end{align*}
for a sufficiently large universal constant $C>0$.
\end{lemma}

\begin{proof}
Let $H=\langle w^\star,X\rangle+\sigma G$. Then $H\sim\mathcal N(0,1+\sigma^2)$, and $Y=-1$ precisely when $H<-t$, so $p_-=\PP(Y=-1)=p_\sigma$. Since $(X, H)$ is jointly Gaussian with $\operatorname{Cov}(X, H) = w^\star$, we have 
\begin{align*}
    \E[\langle w^\star,X\rangle\mid H]
    =
    \frac{
        \operatorname{Cov}(\langle w^\star,X\rangle,H)
    }{\operatorname{Var}(H)}H
    =
    \frac{H}{1+\sigma^2}
\end{align*}
where $\operatorname{Cov}(\langle w^\star,X\rangle,H)=\langle w^\star,\operatorname{Cov}(X,H)\rangle=1$. The component of $X$ orthogonal to $w^\star$ is independent of $H$ and has mean zero. Conditioning further on $H<-t$ and using that $H/\sqrt{1+\sigma^2}$ is standard Gaussian, we obtain
\begin{align*}
    \mu_-
    &=
    \frac{\E[H\mid H<-t]}{1+\sigma^2}w^\star\\
    &=
    \frac{w^\star}{p_\sigma\sqrt{1+\sigma^2}}
    \int_{-\infty}^{-t/\sqrt{1+\sigma^2}}u\phi(u)\,du\\
    &=
    -\frac{\phi(t/\sqrt{1+\sigma^2})}
           {p_\sigma\sqrt{1+\sigma^2}}w^\star
    =
    -\frac{\lambda(t/\sqrt{1+\sigma^2})}
           {\sqrt{1+\sigma^2}}w^\star
\end{align*}
where the integral is evaluated using $\phi'(u)=-u\phi(u)$. From Mills' lower bound, $\lambda(a)\ge a$ for $a>0$, which then gives $\|\mu_-\|_2\ge t/(1+\sigma^2)$.

All that remains is to apply Theorem~\ref{thm:minority_mean_general} with target angle $\tau=c/t$. Since $t\ge1$, $\sin(\tau/2)\ge c/(\pi t)$, and
\begin{align*}
    \|\mu_-\|_2^2\sin^2(\tau/2)
    \ge
    \frac{t^2}{(1+\sigma^2)^2}
    \frac{c^2}{\pi^2t^2}
    =
    \frac{c^2}{\pi^2(1+\sigma^2)^2}
\end{align*}
Substituting this bound and $p_-=p_\sigma$ into Theorem~\ref{thm:minority_mean_general} completes the proof. 
\end{proof}
It remains to reconcile direction recovery with the threshold ERM of Algorithm~\ref{alg:threshold-erm}. 
\begin{theorem}[Excess-risk guarantee under probit noise]
\label{thm:probit-excess-risk}
Assume the probit model with $\sigma>0$ and $t\ge1$. Let $p = \Phi(-t) = \Theta(\epsilon)$, $p_\sigma = \Phi\left(-\frac{t}{\sqrt{1+\sigma^2}}\right)$. There is an algorithm which, given
\begin{align*}
    N
    =
    \widetilde O\left(
        \frac{(d+\log(1/\delta))(1+\sigma^2)^2}{p_\sigma}
        +
        \frac{1}{\epsilon^2}
    \right)
\end{align*}
independent probit samples, returns a halfspace $\widehat h$ such that
\begin{align*}
    \operatorname{err}(\widehat h)-\operatorname{err}(f^\star)
    \le
    \epsilon
\end{align*}
with probability at least $1-\delta$. The algorithm runs in time $\widetilde O(dN)$
\end{theorem}

\begin{proof}
As in proof of Theorem~\ref{thm:standalone-rv-halfspace}, we will first split the samples into two independent sets, and use the first for direction estimation via Algorithm~\ref{alg:rare-tail-direction} and the second for threshold estimation via Algorithm~\ref{alg:threshold-erm}. 

From Lemma~\ref{lem:dir-estimate-under-probit}, applied with confidence parameter $\delta/2$ and a sufficiently small constant $c>0$ such that Lemma~\ref{lem:angle-to-risk-fixed-threshold} holds, we have that 
\begin{align*}
    N_{\rm dir}
    =
    \widetilde O\left(
        \frac{(d+\log(1/\delta))(1+\sigma^2)^2}{p_\sigma}
    \right)
\end{align*}
samples suffice to ensure $\angle(\widehat w,w^\star)\le \frac{c}{t}$ with probability at least $1-\delta/2$.

Conditioned on this event, Theorem~\ref{thm:threshold-erm-excess-risk} applies with confidence parameter $\delta/2$. If the accuracy to which the threshold is to be computed is a sufficiently small constant multiple of $\epsilon$, then 
\begin{align*}
    k
    =
    \widetilde O\left(\frac{1}{\epsilon^2}\right)
\end{align*}
samples ensures
\begin{align*}
    \operatorname{err}(\widehat h)-\operatorname{err}(f^\star)
    \le
    \epsilon
\end{align*}
with probability at least $1-\delta/2$. A union bound completes the proof for the sample complexity, and the runtime argument from the proof of Proposition~\ref{prop:rare-tail-threshold-runtime} applies verbatim. 
\end{proof}

\section{Results for Curriculum Learning}\label{app:curriculum-learning}
 
\subsection{A General Learning Model}\label{subsec:gen-cl-model}
 
Although Section~\ref{sec:curriculum} defines a two-stage curriculum, for the sake of completeness, we record the following general formalization of curriculum learning and its instantiation in the Gaussian halfspace setting. 
\begin{definition}[$r$-stage curriculum learning]
\label{def:r-stage-curriculum}
Let $\mathcal X$ be an instance space, let $\mathcal Y=\{\pm1\}$, and let
$\mathcal H\subseteq\mathcal Y^{\mathcal X}$ be a hypothesis class. Fix an integer $r\ge1$ and stage lengths $n=(n_1,\ldots,n_r)\in\mathbb N^r$.
 
An $r$-stage curriculum family is a collection of stage-indexed distribution families
\begin{align*}
    \mathcal D
    =
    \left(\{D_h^{(j)}:h\in\mathcal H\}\right)_{j=1}^r
\end{align*}
where each $D_h^{(j)}$ is a distribution over $\mathcal X\times\mathcal Y$. We say that the family is task-consistent across stages if, for every $h\in\mathcal H$ and every stage $j\in[r]$, if $(X,Y)\sim D_h^{(j)}$ and $P_{X,h}^{(j)}$ denotes the $X$-marginal of $D_h^{(j)}$, then for $P_{X,h}^{(j)}$-almost every $x\in\mathcal X$,
\begin{align*}
    \PP(Y=h(x)\mid X=x)\ge \frac12
\end{align*}
An $r$-stage curriculum proceeds as follows: an unknown target $h^\star\in\mathcal H$ is fixed. The learner receives $r$ mutually independent stages of i.i.d. labeled samples
\begin{align*}
    S_j=\{(X_i^{(j)},Y_i^{(j)})\}_{i=1}^{n_j}
    \quad\text{i.i.d.}\sim D_{h^\star}^{(j)}
    \qquad\text{for }j\in[r]
\end{align*}
in order. At stage $j$, it uses $S_j$ and the state retained from the preceding stage to compute 
\begin{align*}
    M_j=\mathcal A_j(M_{j-1},S_j)
\end{align*}
where $M_0$ is fixed and the procedures $\mathcal A_j$ may be randomized. This computation is completed before the next stage is revealed, and every subsequent stage can access earlier samples only through the retained state. After the final stage, the learner outputs a classifier $\widehat h=\mathcal B(M_r)$, proper or improper.
 
We refer to the last-stage distribution $D_{h^\star}^{(r)}$ as the target distribution, and to $(D_{h^\star}^{(1)},\ldots,D_{h^\star}^{(r-1)})$ as the curriculum distributions. Performance is measured by excess $0-1$ risk on the target stage, i.e.,
\begin{align*}
    \operatorname{err}_{D_{h^\star}^{(r)}}(\widehat h)
    -
    \operatorname{err}_{D_{h^\star}^{(r)}}(h^\star)
\end{align*}
\end{definition}

\begin{definition}[Gaussian halfspace instantiation]
\label{def:gaussian-instantiation}
Recall the Gaussian halfspace setting. For $(w,t)\in\mathbb S^{d-1}\times\mathbb R$, let
\begin{align*}
    m_{w,t}(x)=\langle w,x\rangle+t,
    \qquad
    f_{w,t}(x)=\operatorname{sign}(m_{w,t}(x))
\end{align*}
Fix an integer $r\ge1$ and choose stage-dependent noise profiles $g_1,\ldots,g_r:[0,\infty)\to[0,1/2]$.

For each stage $j\in[r]$ and each halfspace $f_{w,t}$, let $D_{w,t}^{(j)}$ be the distribution of $(X,Y)$ obtained by drawing $X\sim\mathcal N(0,I_d)$ and then generating $Y$ according to
\begin{align*}
    \PP(Y\ne f_{w,t}(X)\mid X=x)
    =
    g_j(|m_{w,t}(x)|)
\end{align*}
Equivalently,
\begin{align*}
    \mathbb E[Y\mid X=x]
    =
    \left(1-2g_j(|m_{w,t}(x)|)\right)f_{w,t}(x)
\end{align*}
Since $g_j\le1/2$, the halfspace $f_{w,t}$ is Bayes optimal for every stage, and this construction satisfies the task-consistency condition in Definition~\ref{def:r-stage-curriculum}. 
\end{definition}
 
\subsection{A two-stage curriculum learner for $\eta$-RCN targets}
\label{app:curriculum-rcn}
 
Throughout this subsection, the source stage has regularly-varying margin-monotone noise (Definition~\ref{def:regularly-varying-margin-noise}) and the target stage is $\eta$-RCN with known constant $\eta<1/2$. As in the rest of the appendix, the analysis is written under the convention $t>0$ without loss of generality.
 
Recall from Section~\ref{sec:curriculum} that, under $\eta$-RCN, the algorithm of \citet{diakonikolas2023nearoptimalboundslearninggaussian} is expensive only in its first step. Finding a warm start within angle $1/(5t)$ of $w^\star$ from the Average vector takes $\widetilde O(d/\max\{p,\epsilon\}^2)$ samples, whereas turning this warm start into a halfspace with excess risk $\epsilon$ takes only $\widetilde O(d/\epsilon)$. Our curriculum learner (Algorithm~\ref{alg:rare-tail-curriculum-rcn}) therefore computes the warm start from the source samples, where Algorithm~\ref{alg:rare-tail-direction} needs only $\widetilde O(d/p)$ of them, and leaves two tasks to the target samples: fitting the threshold, and shrinking the angle from $O(1/t)$ to $O_\eta(\epsilon\exp(t^2/2))$. This is a reduction by a factor of order $p/\epsilon$, down to the scale at which Lemma~\ref{lem:angle-to-risk-fixed-threshold} bounds the disagreement with $f^\star$ by $O(\epsilon)$. For this second task, we follow the Optimization procedure of \citet[Algorithm~2]{diakonikolas2023nearoptimalboundslearninggaussian}. We first describe this procedure, and then the changes that we make to it.
 
\paragraph{The band update.}
The refinement procedure of
\citet{diakonikolas2023nearoptimalboundslearninggaussian}
uses target samples in a thin band next to its current guess
of the separator. Given a direction $w\in S^{d-1}$, a threshold
guess $t_0$, and a width $\gamma_0>0$, the band is
\begin{align}
    \mathcal E(w,t_0)
    =
    \left\{
        x:-t_0\le\langle w,x\rangle\le-t_0+\gamma_0
    \right\}
    \label{eq:refinement-band}
\end{align}
which lies on the side that the current classifier labels $+1$.
Each sample $(x,y)$ in the band contributes the vector
\begin{align}
    g(w;x,y)
    =
    \frac12
    \left(
        (1-2\eta)\operatorname{sign}(\langle w,x\rangle+t_0)-y
    \right)
    \operatorname{proj}_{w^\perp}(x)
    \label{eq:band-update-vector}
\end{align}
where $\operatorname{proj}_{w^\perp}(x)=x-\langle w,x\rangle w$.
The procedure estimates the conditional mean of these vectors
on the band, takes a small step against this estimate, and
normalizes, keeping $t_0$ fixed.

The reason for this update is that, within the range of the
local descent lemma, the population mean is orthogonal to $w$
and negatively correlated with $w^\star$. We refer to
\citet[Section~2.2]{diakonikolas2023nearoptimalboundslearninggaussian}
for the geometric argument, and establish the corresponding
bound for our threshold guess in
Lemma~\ref{lem:local-pop-descent-band}.
We use the band width
\begin{align*}
    \gamma_0
    =
    \frac{\epsilon}{2(1-2\eta)}
    \exp(t_0^2/2)
\end{align*}
and describe our changes to the procedure below.
 
\paragraph{Our changes.}
In \citet{diakonikolas2023nearoptimalboundslearninggaussian}, the band update is one of three components of the learner (their Algorithm~1). The warm start is the normalized Average vector, which is where the quadratic cost arises. The optimization is then run once for every threshold guess on a grid of spacing $\epsilon^2/(8(1-2\eta)^2)$, which is the accuracy that their descent lemma requires of $t_0$, and a final testing step selects one of the resulting $O(1/\epsilon^2)$ halfspaces. We keep the band update and replace the other two components. The warm start is the output of Algorithm~\ref{alg:rare-tail-direction} on the source samples. For the threshold, we show that the update keeps its direction whenever $|t_0-t|\le\gamma_0$ (Lemma~\ref{lem:local-pop-descent-band}). This is a much weaker requirement, since $\gamma_0\ge\epsilon/(2(1-2\eta))$, and it is met by a guess computed directly from the frequency of the label $-1$, which equals $\eta+(1-2\eta)p$ under $\eta$-RCN (Algorithm~\ref{alg:rcn-band-threshold-from-labels} and Lemma~\ref{lem:rcn-band-threshold-estimation}). The refinement is therefore run only once, which is the source of the runtime gain in Table~\ref{tab:bias-regime-comparison}, and the final threshold is fitted by Algorithm~\ref{alg:threshold-erm} along the refined direction.
 
We also change two details of the procedure itself, at points where we could not verify the analysis of \citet{diakonikolas2023nearoptimalboundslearninggaussian} as stated. The first is the step size. Their schedule starts at $\mu_0=(1-4\rho)\sqrt{2\pi}/(16(1-2\eta))$ with $\rho=0.00098$, which is calibrated for an initial angle of constant order, whereas the descent lemma, theirs as well as ours, holds only within angle $1/(5t)$ of $w^\star$. With this schedule, the first steps can rotate the direction by angles larger than $1/(5t)$ once $t\ge4$, and the induction of \citep[Lemma~2.8]{diakonikolas2023nearoptimalboundslearninggaussian} does not show that the iterates stay in the region where the descent lemma applies. We start the same schedule at the angle of the warm start instead (line~1 of Algorithm~\ref{alg:local-direction-refine}). The induction then goes through as before, and also shows that every iterate stays within angle $1/(5t)$ of $w^\star$.
 
The second is the use of the samples. In \citet{diakonikolas2023nearoptimalboundslearninggaussian}, every iteration uses the same sample, so each iterate depends on the sample that updates it, and the empirical update has to be close to its mean for all directions at once. Their concentration bound \citep[Lemma~2.5]{diakonikolas2023nearoptimalboundslearninggaussian} is proved for a fixed direction and extended by a union bound over a cover, but do not address directions outside the cover. It is also stated only for the angles covered by the descent lemma, while the induction applies it to iterates that are already closer to $w^\star$. We instead split the sample into one batch per iteration, so that each iterate is independent of the batch that updates it, and prove a bound for a fixed direction that holds at every angle (Lemma~\ref{lem:local-update-concentration}). Since there are $O(\log(1/\epsilon))$ iterations, the refinement stage still uses $\widetilde O_\eta((d+\log(1/\delta))/\epsilon)$ target samples.
 
\begin{algorithm}[t]
\caption{\textsc{RCN band-Scale Threshold from Labels}}
\label{alg:rcn-band-threshold-from-labels}
\begin{algorithmic}[1]
\Require Target RCN samples $S=\{(X_i,Y_i)\}_{i=1}^k$, known noise rate $\eta<1/2$
\Ensure A threshold estimate $\widehat t$
\State $N_-\gets \sum_{i=1}^k\mathbf 1\{Y_i=-1\}$, \quad
       $N_+\gets \sum_{i=1}^k\mathbf 1\{Y_i=+1\}$
\State $\widehat y_{\rm rt}\gets -1$ if $N_-\le N_+$, and $\widehat y_{\rm rt}\gets +1$ otherwise
\State $\widehat q_{\rm rt}\gets k^{-1}\sum_{i=1}^k\mathbf 1\{Y_i=\widehat y_{\rm rt}\}$
\State $\widehat p_{\rm raw}\gets (\widehat q_{\rm rt}-\eta)/(1-2\eta)$
\State $\widehat p\gets \Pi_{[0,1/2]}(\widehat p_{\rm raw})$ \hfill (clip to the valid bias range)
\If{$\widehat p=0$}
    \State $\widehat a\leftarrow1$
\Else
    \State $\widehat a\gets \Phi^{-1}(1-\widehat p)$
\EndIf
\State $\widehat t\gets -\widehat y_{\rm rt}\widehat a$
\State \Return $\widehat t$
\end{algorithmic}
\end{algorithm}
 
\begin{algorithm}[t]
\caption{Local direction refinement, adapted from \citet[Algorithm~2]{diakonikolas2023nearoptimalboundslearninggaussian}}
\label{alg:local-direction-refine}
\begin{algorithmic}[1]
\Require Initial direction $w_0\in S^{d-1}$, threshold estimate $t_0$, band width $\gamma_0$, known noise rate $\eta<1/2$, target RCN samples $S=\{(X_i,Y_i)\}_{i=1}^{N_2}$
\Ensure A refined direction
\textbf{if} $t_0<1/2$ \textbf{then return} $w_0$
\State Set $\rho_{\rm loc}\gets 0.00098$, $\bar\theta\gets 1/(6t_0)$, and
\begin{align*}
    \mu_0\gets \frac{(1-4\rho_{\rm loc})\sqrt{2\pi}\,\sin(\bar\theta/2)}{16(1-2\eta)},
    \qquad
    K\gets
    \left\lfloor
        \frac{\log\left(3\sin(\bar\theta/2)/(16\gamma_0)\right)}
             {\log\left(1/(1-\rho_{\rm loc})\right)}
    \right\rfloor
\end{align*}
\State \textbf{if} $K<0$ \textbf{then return} $w_0$
\State Set
\begin{align*}
    P(t_0,\gamma_0)
    \gets
    \PP_{Z\sim\mathcal N(0,1)}
    \left[-t_0\le Z\le -t_0+\gamma_0\right]
\end{align*}
\State Split $S$ into $K+1$ batches $S^{(0)},\ldots,S^{(K)}$ of equal size $m=\lfloor N_2/(K+1)\rfloor$
\For{$k=0,1,\ldots,K$}
    \State Define the current band
    \begin{align*}
        \mathcal E_k
        =
        \left\{
            x:
            -t_0\le \langle w_k,x\rangle\le -t_0+\gamma_0
        \right\}
    \end{align*}
    \State For each $(X_i,Y_i)\in S^{(k)}$, set
    \begin{align*}
        G_k(X_i,Y_i)
        =
        \frac12
        \left(
            (1-2\eta)\operatorname{sign}(\langle w_k,X_i\rangle+t_0)
            -
            Y_i
        \right)
        \operatorname{proj}_{w_k^\perp}(X_i)
    \end{align*}
    \State Form the empirical localized update from the batch $S^{(k)}$
    \begin{align*}
        \widehat G_k
        =
        \frac1{m}
        \sum_{(X_i,Y_i)\in S^{(k)}}
        G_k(X_i,Y_i)
        \frac{\mathbf 1\{X_i\in \mathcal E_k\}}
        {P(t_0,\gamma_0)}
    \end{align*}
    \State $\mu_k\gets \mu_0(1-\rho_{\rm loc})^k$
    \State Update
    \begin{align*}
        w_{k+1}
        \gets
        \frac{w_k-\mu_k\widehat G_k}{\|w_k-\mu_k\widehat G_k\|_2}
    \end{align*}
\EndFor
\State \Return $w_{K+1}$
\end{algorithmic}
\end{algorithm}
 
\begin{algorithm}[t]
\caption{\textsc{Rare-Tail Curriculum Learner for $\eta$-RCN Targets}}
\label{alg:rare-tail-curriculum-rcn}
\begin{algorithmic}[1]
\Require Source samples $S_1$, target RCN samples $S_2$, accuracy $\epsilon$, failure probability $\delta$, known noise rate $\eta<1/2$
\Ensure A halfspace $\widehat h$
\State Split $S_2$ into independent parts $S_{2,\rm band}$, $S_{2,\rm ref}$, and $S_{2,\rm thr}$
\State $(\widehat w_0,\widehat y_{\rm rt})\gets \textsc{Rare-Tail Direction Estimator}(S_1)$
\State $\widehat t_0\gets \textsc{RCN band-Scale Threshold from Labels}(S_{2,\rm band},\eta)$
\State Set
\begin{align*}
    \widehat\gamma
    \gets
    \frac{\epsilon}{2(1-2\eta)}
    \exp(\widehat t_0^2/2)
\end{align*}
\State $\widehat w_{\rm ref}\gets \textsc{Local Direction Refinement}(\widehat w_0,\widehat t_0,\widehat\gamma,\eta,S_{2,\rm ref})$
\State $\widehat t_{\rm final}\gets \textsc{Threshold-ERM}(S_{2,\rm thr},\widehat w_{\rm ref})$
\State \Return $\widehat h(x)=\operatorname{sign}(\langle \widehat w_{\rm ref},x\rangle+\widehat t_{\rm final})$
\end{algorithmic}
\end{algorithm}
 
We begin the analysis with the threshold guess. Since the label $-1$ has frequency $\eta+(1-2\eta)p$ under $\eta$-RCN, estimating this frequency to accuracy of order $\epsilon$ determines $p$, and hence $t$, up to a fraction of the band width.
 
\begin{lemma}[Band-scale threshold estimation from target RCN labels]
\label{lem:rcn-band-threshold-estimation}
Assume the target stage is $\eta$-RCN with known constant $\eta<1/2$. Suppose $t\ge1$ and
$p=\Phi(-t)\ge C_\eta\epsilon$ for a sufficiently large constant $C_\eta>0$ depending only on
$\eta$. Define
\begin{align*}
    \gamma_\star
    =
    \frac{\epsilon}{2(1-2\eta)}
    \exp(t^2/2)
\end{align*}
For every constant $a\in(0,1/2]$, Algorithm~\ref{alg:rcn-band-threshold-from-labels}
returns $\widehat t$ satisfying
\begin{align*}
    |\widehat t-t|\le a\gamma_\star
\end{align*}
with probability at least $1-\delta$, using
\begin{align*}
    k
    =
    O_\eta\left(\frac{1}{a^2\epsilon^2}\log(1/\delta)\right)
\end{align*}
target RCN samples.
\end{lemma}
 
\begin{proof}
Since $t>0$, the clean rare label is $-1$. Under $\eta$-RCN,
\begin{align*}
    q
    =
    \PP(Y=-1)
    =
    \eta+(1-2\eta)p
\end{align*}
and we write $\widehat q=N_-/k$ for the empirical frequency of the label $-1$.
 
We first record that the band is narrow in the regime $p\ge C_\eta\epsilon$. Since $\exp(t^2/2)=1/(\sqrt{2\pi}\phi(t))$ and $\Phi(-t)\le\phi(t)/t$,
\begin{align}
    t\gamma_\star
    =
    \frac{t\epsilon}{2(1-2\eta)\sqrt{2\pi}\phi(t)}
    \le
    \frac{\epsilon}{2\sqrt{2\pi}(1-2\eta)p}
    \le
    \frac{1}{2\sqrt{2\pi}(1-2\eta)C_\eta}
    \le
    \frac1{40}
    \label{eq:band-scale}
\end{align}
where the last inequality holds once $C_\eta\ge 8/(1-2\eta)$. Since $t\ge1$, this also gives $\gamma_\star\le1/40$.
 
\emph{Identifying the rare label.} Since $t\ge1$, we have $p\le\Phi(-1)<1/4$, which gives
\begin{align*}
    \frac12-q
    =
    (1-2\eta)\left(\frac12-p\right)
    \ge
    \frac{1-2\eta}{4}
\end{align*}
Algorithm~\ref{alg:rcn-band-threshold-from-labels} sets $\widehat y_{\rm rt}=-1$ unless $N_->N_+$, that is, unless $\widehat q>1/2$. By Hoeffding's inequality,
\begin{align*}
    \PP(\widehat y_{\rm rt}\ne-1)
    \le
    \PP\left(
        \widehat q-q
        >
        \frac12-q
    \right)
    \le
    \exp\left(
        -2k\left(\frac12-q\right)^2
    \right)
    \le
    \exp\left(
        -\frac{k(1-2\eta)^2}{8}
    \right)
\end{align*}
which is at most $\delta/2$ once $k\ge 8\log(2/\delta)/(1-2\eta)^2$.
 
\emph{From the label frequency to the threshold.} Let $\tau=a\gamma_\star\phi(t+a\gamma_\star)$. We show that, on the event where $\widehat y_{\rm rt}=-1$ and $|\widehat q-q|\le(1-2\eta)\tau$, the output satisfies $|\widehat t-t|\le a\gamma_\star$. On this event, $\widehat q_{\rm rt}=\widehat q$, and therefore
\begin{align*}
    |\widehat p_{\rm raw}-p|
    =
    \frac{|\widehat q-q|}{1-2\eta}
    \le
    \tau
\end{align*}
Since $p\in[0,1/2]$, clipping $\widehat p_{\rm raw}$ to this interval cannot increase its distance to $p$, and $|\widehat p-p|\le\tau$. Next, $\phi$ is decreasing on $[0,\infty)$ and $t-a\gamma_\star\ge0$ by eq.~\eqref{eq:band-scale}, hence
\begin{align*}
    \Phi(-(t-a\gamma_\star))-p
    &=
    \int_{t-a\gamma_\star}^{t}\phi(u)\,du
    \ge
    a\gamma_\star\phi(t)
    \ge
    \tau \\
    p-\Phi(-(t+a\gamma_\star))
    &=
    \int_t^{t+a\gamma_\star}\phi(u)\,du
    \ge
    a\gamma_\star\phi(t+a\gamma_\star)
    =
    \tau
\end{align*}
Combining these bounds with $|\widehat p-p|\le\tau$ gives
\begin{align*}
    0
    <
    \Phi(-(t+a\gamma_\star))
    \le
    \widehat p
    \le
    \Phi(-(t-a\gamma_\star))
\end{align*}
In particular $\widehat p>0$, so the algorithm sets $\widehat a=\Phi^{-1}(1-\widehat p)$, which is the same as $\Phi(-\widehat a)=\widehat p$. Since $\Phi$ is increasing, this gives $t-a\gamma_\star\le\widehat a\le t+a\gamma_\star$. Finally, $\widehat y_{\rm rt}=-1$ gives $\widehat t=\widehat a$, and $|\widehat t-t|\le a\gamma_\star$.
 
\emph{Sample size.} It remains to bound the probability that $|\widehat q-q|>(1-2\eta)\tau$. By definition of $\gamma_\star$,
\begin{align*}
    \gamma_\star\phi(t)
    =
    \frac{\epsilon}{2(1-2\eta)}
    \exp(t^2/2)
    \cdot
    \frac{1}{\sqrt{2\pi}}\exp(-t^2/2)
    =
    \frac{\epsilon}{2(1-2\eta)\sqrt{2\pi}}
\end{align*}
and by eq.~\eqref{eq:band-scale} and $a\le1/2$,
\begin{align*}
    \frac{\phi(t+a\gamma_\star)}{\phi(t)}
    =
    \exp\left(-at\gamma_\star-\frac{a^2\gamma_\star^2}{2}\right)
    \ge
    \exp\left(-\frac1{40}\right)
    \ge
    \frac12
\end{align*}
Therefore $(1-2\eta)\tau\ge a\epsilon/(4\sqrt{2\pi})$, and Hoeffding's inequality gives
\begin{align*}
    \PP\left(
        |\widehat q-q|>(1-2\eta)\tau
    \right)
    \le
    2\exp\left(-2k(1-2\eta)^2\tau^2\right)
    \le
    2\exp\left(-\frac{ka^2\epsilon^2}{16\pi}\right)
\end{align*}
which is at most $\delta/2$ once $k\ge 16\pi\log(4/\delta)/(a^2\epsilon^2)$. A union bound over the two events completes the proof.
\end{proof}
 
The next lemma is the population statement behind the band update, and our counterpart to \citep[Lemmas~2.3 and~2.4]{diakonikolas2023nearoptimalboundslearninggaussian}. Their threshold-accuracy assumption, written in our notation, is
\begin{align*}
    |t_0-t|\le\frac{\epsilon^2}{8(1-2\eta)^2}
\end{align*}
where we have substituted their $\epsilon'=\epsilon/(1-2\eta)$. We show that the population descent bound holds under the weaker requirement
\begin{align*}
    |t_0-t|\le\gamma_0,
\end{align*}
provided $8\gamma_0\le\theta\le1/(5t)$.
\begin{lemma}[Population descent under a band-accurate threshold guess]
\label{lem:local-pop-descent-band}
Assume the target stage is $\eta$-RCN with known constant $\eta<1/2$, and suppose $t\ge1$. Let $w\in S^{d-1}$, $t_0\in\mathbb R$, and $\gamma_0>0$, let $\mathcal E(w,t_0)$ be the band of eq.~\eqref{eq:refinement-band} and $g(w;x,y)$ the vector of eq.~\eqref{eq:band-update-vector}, and define the population localized update
\begin{align*}
    G(w)=\mathbb E[g(w;X,Y)\mid X\in \mathcal E(w,t_0)]
\end{align*}
Then
\begin{align*}
    \|G(w)\|_2\le C_1=\frac{1-2\eta}{\sqrt{2\pi}}
\end{align*}
Moreover, if $\theta=\angle(w,w^\star)$ satisfies
\begin{align*}
    8\gamma_0\le\theta\le \frac{1}{5t}
    \qquad\text{and}\qquad
    |t_0-t|\le \gamma_0
\end{align*}
then
\begin{align*}
    G(w)\cdot w^\star
    \le
    -\phi(7/10)(1-2\eta)\sin\theta
    \le
    -\frac{C_1}{2}\sin\theta
\end{align*}
\end{lemma}
 
\begin{proof}
On the band event $X\in\mathcal E(w,t_0)$, we have
$\operatorname{sign}(\langle w,X\rangle+t_0)=+1$. Also, under $\eta$-RCN,
\begin{align*}
    \mathbb E[Y\mid X]=(1-2\eta)f^\star(X)
\end{align*}
Therefore
\begin{align*}
    G(w)
    &=
    \frac12
    \mathbb E\left[
        \left(
            (1-2\eta)-(1-2\eta)f^\star(X)
        \right)
        \operatorname{proj}_{w^\perp}(X)
        \mid X\in\mathcal E(w,t_0)
    \right] \\
    &=
    (1-2\eta)
    \mathbb E\left[
        \mathbf 1\{f^\star(X)=-1\}
        \operatorname{proj}_{w^\perp}(X)
        \mid X\in\mathcal E(w,t_0)
    \right]
\end{align*}
If $w=\pm w^\star$, then both $f^\star(X)$ and the band event are determined by $\langle w,X\rangle$, which is independent of $\operatorname{proj}_{w^\perp}(X)$, so $G(w)=0$ and the norm bound is trivial. Otherwise $\theta\in(0,\pi)$, and by rotational invariance we may assume $w=e_1$ and
\begin{align*}
    w^\star=\cos\theta\,e_1+\sin\theta\,e_2
\end{align*}
Let $Z=\langle w,X\rangle$ and $U=\langle e_2,X\rangle$. Then $Z,U$ are independent standard
Gaussians. The event $f^\star(X)=-1$ is
\begin{align*}
    \cos\theta\,Z+\sin\theta\,U+t<0
\end{align*}
or equivalently $U<a(Z)$, where
\begin{align*}
    a(z)=\frac{-t-\cos\theta\,z}{\sin\theta}
\end{align*}
The coordinates of $X$ orthogonal to $\operatorname{span}\{e_1,e_2\}$ are independent of $(Z,U)$ and have mean zero, so
$G(w)$ is collinear with $e_2$. Since $\mathbb E[U\mathbf 1\{U<a\}]=-\phi(a)$ for every $a\in\mathbb R$,
we get
\begin{align*}
    G(w)\cdot e_2
    &=
    (1-2\eta)
    \mathbb E[
        U\mathbf 1\{U<a(Z)\}
        \mid
        Z\in[-t_0,-t_0+\gamma_0]
    ] \\
    &=
    -(1-2\eta)
    \mathbb E[
        \phi(a(Z))
        \mid
        Z\in[-t_0,-t_0+\gamma_0]
    ]
\end{align*}
Since $0<\phi\le\phi(0)=1/\sqrt{2\pi}$, this proves the first claim,
\begin{align*}
    \|G(w)\|_2
    =
    |G(w)\cdot e_2|
    \le
    \frac{1-2\eta}{\sqrt{2\pi}}
\end{align*}
Moreover, $w^\star\cdot e_2=\sin\theta$, and
\begin{align*}
    G(w)\cdot w^\star
    =
    -(1-2\eta)\sin\theta\,
    \mathbb E[
        \phi(a(Z))
        \mid
        Z\in[-t_0,-t_0+\gamma_0]
    ]
\end{align*}
For the second claim, it remains to lower bound $\phi(a(z))$ uniformly over the band. Write $z=-t_0+s$ with
$s\in[0,\gamma_0]$ and $\Delta=t_0-t$. Then
\begin{align*}
    -t-\cos\theta\,z
    =
    t(\cos\theta-1)+\cos\theta\,\Delta-\cos\theta\,s
\end{align*}
which then gives
\begin{align*}
    |a(z)|
    \le
    \frac{t(1-\cos\theta)}{\sin\theta}
    +
    \frac{|\Delta|}{\sin\theta}
    +
    \frac{s}{\sin\theta}
\end{align*}
Since $t\ge1$, we have $\theta\le1/(5t)\le1$. Using $1-\cos\theta\le \theta^2/2$ and $\sin\theta\ge \theta/2$ for $\theta\in(0,1]$, together
with $\theta\le 1/(5t)$, $\theta\ge8\gamma_0$, $|\Delta|\le \gamma_0$, and $s\le\gamma_0$, we have
\begin{align*}
    |a(z)|
    \le
    t\theta
    +
    2\frac{\gamma_0}{\theta}
    +
    2\frac{\gamma_0}{\theta}
    \le
    \frac15+\frac14+\frac14
    =
    \frac{7}{10}
\end{align*}
Hence $\phi(a(z))\ge \phi(7/10)$ throughout the band, and
\begin{align*}
    G(w)\cdot w^\star
    \le
    -\phi(7/10)(1-2\eta)\sin\theta
\end{align*}
Finally, $\phi(7/10)=e^{-49/200}/\sqrt{2\pi}\ge 1/(2\sqrt{2\pi})$, which gives the bound in terms of $C_1$.
\end{proof}
 
The empirical update of Algorithm~\ref{alg:local-direction-refine} averages the vectors of eq.~\eqref{eq:band-update-vector} over the samples of one batch that fall in the band. For a direction that does not depend on the batch, this average is close to $G(w)$, whatever the angle between $w$ and $w^\star$.
 
\begin{lemma}[Concentration of the localized update for a fixed direction]
\label{lem:local-update-concentration}
Assume the target stage is $\eta$-RCN with known constant $\eta<1/2$. Fix $w\in S^{d-1}$, $t_0\in\mathbb R$, and $\gamma_0>0$, and let
\begin{align*}
    P=\PP_{Z\sim\mathcal N(0,1)}\left[-t_0\le Z\le -t_0+\gamma_0\right]
\end{align*}
be the Gaussian mass of the band $\mathcal E(w,t_0)$. Let $(X_1,Y_1),\ldots,(X_m,Y_m)$ be i.i.d. target RCN samples, and define
\begin{align*}
    \widehat G(w)
    =
    \frac{1}{mP}
    \sum_{i=1}^m
    g(w;X_i,Y_i)\,
    \mathbf 1\{X_i\in\mathcal E(w,t_0)\}
\end{align*}
For every $\delta\in(0,1)$, if
\begin{align*}
    m
    \ge
    \frac{2048\pi}{(1-2\eta)^2P}
    \left(d\log 5+\log(2/\delta)\right)
\end{align*}
then, with probability at least $1-\delta$,
\begin{align*}
    \|\widehat G(w)-G(w)\|_2
    \le
    \frac{C_1}{4}
    =
    \frac{1-2\eta}{4\sqrt{2\pi}}
\end{align*}
\end{lemma}
 
\begin{proof}
Let $B_i=\mathbf 1\{X_i\in\mathcal E(w,t_0)\}$, so that $\PP(B_i=1)=P$. On the band, $\operatorname{sign}(\langle w,X_i\rangle+t_0)=+1$, and therefore
\begin{align*}
    B_i\,g(w;X_i,Y_i)
    =
    B_i\,c_i\operatorname{proj}_{w^\perp}(X_i),
    \qquad
    c_i=\frac12\left((1-2\eta)-Y_i\right)\in\{-\eta,1-\eta\}
\end{align*}
In particular, $|c_i|\le1$.
 
Fix a unit vector $v\in S^{d-1}$, and let $h_i=\langle v,\operatorname{proj}_{w^\perp}(X_i)\rangle=\langle \operatorname{proj}_{w^\perp}(v),X_i\rangle$. This is a centered Gaussian with variance $\sigma_v^2=\|\operatorname{proj}_{w^\perp}(v)\|_2^2\le1$, and it is independent of $\langle w,X_i\rangle$, and, therefore, of $B_i$. Let
\begin{align*}
    \xi_i=B_ic_ih_i=\langle v,B_i\,g(w;X_i,Y_i)\rangle
\end{align*}
so that $\langle v,\widehat G(w)\rangle=(mP)^{-1}\sum_{i=1}^m\xi_i$, and $\mathbb E[\xi_i]=P\langle v,G(w)\rangle$ by the definition of $G(w)$. We bound the moment generating function of $\xi_i$. For every $x\in\mathbb R$, Taylor's theorem gives $e^x\le1+x+\frac{x^2}{2}e^{|x|}$. We apply this with $x=\lambda\xi_i$ for $|\lambda|\le1$. Since $|\xi_i|\le B_i|h_i|$, and since $B_i$ and $h_i$ are independent,
\begin{align*}
    \mathbb E\left[e^{\lambda\xi_i}\right]
    \le
    1+\lambda\,\mathbb E[\xi_i]
    +\frac{\lambda^2}{2}\,\mathbb E\left[B_ih_i^2e^{|h_i|}\right]
    =
    1+\lambda P\langle v,G(w)\rangle
    +\frac{\lambda^2P}{2}\,\mathbb E\left[h_i^2e^{|h_i|}\right]
\end{align*}
A centered Gaussian $h$ with variance $\sigma^2\le1$ satisfies $\mathbb E[h^2e^{h}]=(\sigma^2+\sigma^4)e^{\sigma^2/2}\le2\sqrt e$, and $e^{|h|}\le e^{h}+e^{-h}$, so the last expectation is at most $4\sqrt e\le8$. Using $1+x\le e^x$, we obtain
\begin{align*}
    \mathbb E\left[\exp\left(\lambda(\xi_i-\mathbb E[\xi_i])\right)\right]
    \le
    \exp\left(4P\lambda^2\right)
    \qquad\text{for all }|\lambda|\le1
\end{align*}
The samples are independent, so for every $s\in[0,8]$, the Chernoff bound with $\lambda=s/8$ gives
\begin{align*}
    \PP\left(
        \langle v,\widehat G(w)-G(w)\rangle\ge s
    \right)
    &=
    \PP\left(
        \sum_{i=1}^m(\xi_i-\mathbb E[\xi_i])\ge mPs
    \right) \\
    &\le
    \exp\left(-\lambda mPs+4mP\lambda^2\right)
    =
    \exp\left(-\frac{mPs^2}{16}\right)
\end{align*}
Applying the same bound to $-v$ shows that $|\langle v,\widehat G(w)-G(w)\rangle|\ge s$ with probability at most $2\exp(-mPs^2/16)$.
 
Let $\mathcal N_{1/2}$ be a $1/2$-net of $S^{d-1}$ with $|\mathcal N_{1/2}|\le5^d$, as in Lemma~\ref{lem:eps_net_sphere}, so that every $u\in\mathbb R^d$ satisfies
\begin{align*}
    \|u\|_2\le2\max_{v\in\mathcal N_{1/2}}|\langle v,u\rangle|
\end{align*}
Taking $s=C_1/8$ and a union bound over the net,
\begin{align*}
    \PP\left(
        \|\widehat G(w)-G(w)\|_2>\frac{C_1}{4}
    \right)
    \le
    2\cdot5^d\exp\left(-\frac{mPC_1^2}{1024}\right)
    =
    2\cdot5^d\exp\left(-\frac{mP(1-2\eta)^2}{2048\pi}\right)
\end{align*}
which is at most $\delta$ under the stated lower bound on $m$.
\end{proof}
 
We can now analyze Algorithm~\ref{alg:local-direction-refine}. The proof follows the induction of \citep[Lemma~2.8]{diakonikolas2023nearoptimalboundslearninggaussian}, started at the angle of the warm start.
 
\begin{theorem}[Local refinement from a band-accurate threshold guess]
\label{app:thm:band-scale-local-refinement}
Assume the target stage is $\eta$-RCN with known constant $\eta<1/2$. Suppose $t\ge1$ and
$p=\Phi(-t)\ge C_\eta\epsilon$. Define
\begin{align*}
    \gamma_\star
    =
    \frac{\epsilon}{2(1-2\eta)}
    \exp(t^2/2)
\end{align*}
Assume the initialization satisfies
\begin{align*}
    \angle(w_0,w^\star)\le \frac{1}{7t},
    \qquad
    |t_0-t|\le a\gamma_\star
\end{align*}
for a constant $a\in(0,1/2]$. Set
\begin{align*}
    \gamma_0
    =
    \frac{\epsilon}{2(1-2\eta)}
    \exp(t_0^2/2)
\end{align*}
Then Algorithm~\ref{alg:local-direction-refine}, initialized at $(w_0,t_0,\gamma_0)$ and run on
\begin{align*}
    N_2
    =
    \widetilde O_\eta\left(\frac{d+\log(1/\delta)}{\epsilon}\right)
\end{align*}
target RCN samples that are independent of $(w_0,t_0)$, returns a direction $\widehat w_{\rm ref}$ satisfying
\begin{align*}
    \angle(\widehat w_{\rm ref},w^\star)
    \le
    11\gamma_0
    \le
    12\gamma_\star
    =
    \frac{6}{1-2\eta}\,\epsilon\exp(t^2/2)
\end{align*}
with probability at least $1-\delta$. The algorithm runs in time $O(dN_2)$.
\end{theorem}
 
\begin{proof}
\emph{The band width.} By eq.~\eqref{eq:band-scale}, $t\gamma_\star\le1/40$ and $\gamma_\star\le1/40$, so $|t_0-t|\le a\gamma_\star\le1/80$. Hence
\begin{align*}
    |t_0^2-t^2|
    =
    |t_0-t|\,|t_0+t|
    \le
    a\gamma_\star(2t+a\gamma_\star)
    \le
    \frac1{30}
\end{align*}
and since $\gamma_0/\gamma_\star=\exp((t_0^2-t^2)/2)$, we get $e^{-1/60}\gamma_\star\le\gamma_0\le e^{1/60}\gamma_\star$. In particular,
\begin{align*}
    |t_0-t|\le\frac{\gamma_\star}{2}\le\gamma_0
    \qquad\text{and}\qquad
    11\gamma_0\le12\gamma_\star
\end{align*}
Moreover, $|t_0-t|\le1/80\le t/6$ gives $5t/6\le t_0\le7t/6$, so the angle $\bar\theta=1/(6t_0)$ of Algorithm~\ref{alg:local-direction-refine} satisfies
\begin{align*}
    \frac1{7t}\le\bar\theta\le\frac1{5t}
\end{align*}
 
\emph{The empirical updates.} Write $\rho=\rho_{\rm loc}$, $\theta_k=\angle(w_k,w^\star)$, and $\psi_k=\sin(\bar\theta/2)(1-\rho)^k$, so that $\mu_k=(1-4\rho)\psi_k/(16C_1)$ with $C_1=(1-2\eta)/\sqrt{2\pi}$. Suppose first that $K\ge0$. The iterate $w_k$ is a function of $w_0$ and of the batches $S^{(0)},\ldots,S^{(k-1)}$, and is therefore independent of the batch $S^{(k)}$. Conditioned on earlier batches, we can apply Lemma~\ref{lem:local-update-concentration} to the fixed direction $w_k$ and the batch $S^{(k)}$, with confidence parameter $\delta/(K+1)$. A union bound over $k=0,\ldots,K$ shows that, with probability at least $1-\delta$,
\begin{align}
    \|\widehat G_k-G(w_k)\|_2\le\frac{C_1}{4}
    \qquad\text{for every }k\in\{0,\ldots,K\}
    \label{eq:update-tracking}
\end{align}
provided that each batch has size
\begin{align}
    m
    \ge
    \frac{2048\pi}{(1-2\eta)^2P(t_0,\gamma_0)}
    \left(d\log5+\log\frac{2(K+1)}{\delta}\right)
    \label{eq:batch-size}
\end{align}
We verify eq.~\eqref{eq:batch-size} at the end of the proof, and work on the event of eq.~\eqref{eq:update-tracking} until then. On this event, the norm bound of Lemma~\ref{lem:local-pop-descent-band}, which holds for every direction, gives
\begin{align}
    \|\widehat G_k\|_2
    \le
    \|G(w_k)\|_2+\frac{C_1}{4}
    \le
    2C_1
    \qquad\text{for every }k\in\{0,\ldots,K\}
    \label{eq:update-norm}
\end{align}
 
\emph{The induction.} We show that $\sin(\theta_k/2)\le\psi_k$ for $k=0,\ldots,K+1$. The base case holds because $\theta_0\le1/(7t)\le\bar\theta$. Suppose that $\sin(\theta_k/2)\le\psi_k$ for some $k\le K$. Every summand of $\widehat G_k$ is a multiple of $\operatorname{proj}_{w_k^\perp}(X_i)$, so $\widehat G_k$ is orthogonal to $w_k$, and $\|w_k-\mu_k\widehat G_k\|_2\ge1$. The normalization in the update is therefore the Euclidean projection onto the unit ball, which is non-expansive and fixes both $w_k$ and $w^\star$. Consequently,
\begin{align}
    \|w_{k+1}-w_k\|_2\le\mu_k\|\widehat G_k\|_2
    \qquad\text{and}\qquad
    \|w_{k+1}-w^\star\|_2\le\|w_k-\mu_k\widehat G_k-w^\star\|_2
    \label{eq:nonexpansive}
\end{align}
We also recall that $\|u-w^\star\|_2=2\sin(\angle(u,w^\star)/2)$ for every unit vector $u$. We distinguish two cases.
 
Suppose first that $\sin(\theta_k/2)\le\frac34\psi_k$. By the triangle inequality, eq.~\eqref{eq:nonexpansive} and eq.~\eqref{eq:update-norm},
\begin{align*}
    2\sin(\theta_{k+1}/2)
    \le
    2\sin(\theta_k/2)+2\mu_kC_1
    \le
    \frac32\psi_k+\frac{1-4\rho}{8}\psi_k
\end{align*}
and therefore $\sin(\theta_{k+1}/2)\le\frac{13-4\rho}{16}\psi_k\le(1-\rho)\psi_k=\psi_{k+1}$.
 
Suppose now that $\sin(\theta_k/2)>\frac34\psi_k$. The choice of $K$ in Algorithm~\ref{alg:local-direction-refine} gives $\frac34\psi_k\ge\frac34\psi_K\ge4\gamma_0$, so that $\theta_k\ge2\sin(\theta_k/2)>8\gamma_0$. On the other hand, $\sin(\theta_k/2)\le\psi_k\le\sin(\bar\theta/2)$ gives $\theta_k\le\bar\theta\le1/(5t)$. Together with $|t_0-t|\le\gamma_0$, this places $w_k$ in the range of Lemma~\ref{lem:local-pop-descent-band}, and thus
\begin{align*}
    G(w_k)\cdot w^\star\le-\frac{C_1}{2}\sin\theta_k
\end{align*}
The vector $\widehat G_k-G(w_k)$ is orthogonal to $w_k$, and the component of $w^\star$ orthogonal to $w_k$ has norm $\sin\theta_k$. Eq.~\eqref{eq:update-tracking} therefore gives $|(\widehat G_k-G(w_k))\cdot w^\star|\le\frac{C_1}{4}\sin\theta_k$, and hence
\begin{align*}
    \widehat G_k\cdot w^\star\le-\frac{C_1}{4}\sin\theta_k
\end{align*}
Expanding the square on the right-hand side of eq.~\eqref{eq:nonexpansive}, and using $\widehat G_k\cdot w_k=0$ and eq.~\eqref{eq:update-norm}, we obtain the one-step inequality of \citep[Claim~2.7]{diakonikolas2023nearoptimalboundslearninggaussian},
\begin{align*}
    4\sin^2(\theta_{k+1}/2)
    &\le
    \|w_k-w^\star\|_2^2
    +2\mu_k\,\widehat G_k\cdot w^\star
    +\mu_k^2\|\widehat G_k\|_2^2 \\
    &\le
    4\sin^2(\theta_k/2)
    -\frac{C_1}{2}\mu_k\sin\theta_k
    +4C_1^2\mu_k^2
\end{align*}
Since $\theta_k\le1/5$, we have $\sin\theta_k\ge\sin(\theta_k/2)>\frac34\psi_k$. Substituting this bound, the induction hypothesis, and $\mu_k=(1-4\rho)\psi_k/(16C_1)$ gives
\begin{align*}
    4\sin^2(\theta_{k+1}/2)
    \le
    4\psi_k^2
    -\frac{3(1-4\rho)}{128}\psi_k^2
    +\frac{(1-4\rho)^2}{64}\psi_k^2
    =
    4\psi_k^2\left(1-\frac{(1-4\rho)(1+8\rho)}{512}\right)
\end{align*}
The value $\rho=0.00098$ satisfies $32\rho^2+1020\rho\le1$, which is equivalent to $(1-4\rho)(1+8\rho)/512\ge2\rho$. Hence
\begin{align*}
    \sin(\theta_{k+1}/2)
    \le
    \psi_k\sqrt{1-2\rho}
    \le
    (1-\rho)\psi_k
    =
    \psi_{k+1}
\end{align*}
which completes the induction.
 
\emph{The final angle.} Let $\widehat\theta=\angle(\widehat w_{\rm ref},w^\star)$. If $K\ge0$, the algorithm returns $\widehat w_{\rm ref}=w_{K+1}$, and the induction gives $\sin(\widehat\theta/2)\le\psi_{K+1}$, where $\psi_{K+1}<16\gamma_0/3$ by the choice of $K$. If $K<0$, the algorithm returns $\widehat w_{\rm ref}=w_0$, and $\sin(\widehat\theta/2)\le\sin(\bar\theta/2)<16\gamma_0/3$, again by the definition of $K$. In both cases, $\sin(\widehat\theta/2)\le\sin(\bar\theta/2)$, so that $\widehat\theta/2\le\bar\theta/2\le1/10$. Since $x\le\frac{201}{200}\sin x$ for $x\in[0,1/10]$, we conclude that
\begin{align*}
    \angle(\widehat w_{\rm ref},w^\star)
    \le
    \frac{201}{100}\cdot\frac{16}{3}\gamma_0
    \le
    11\gamma_0
    \le
    12\gamma_\star
\end{align*}
 
\emph{Sample size and runtime.} It remains to verify eq.~\eqref{eq:batch-size}. We have $\gamma_0\le2\gamma_\star\le1/20$ and $t_0\ge5/6$, so the interval $[t_0-\gamma_0,t_0]$ lies in $[0,\infty)$, where $\phi$ is decreasing. Hence
\begin{align*}
    P(t_0,\gamma_0)
    =
    \int_{t_0-\gamma_0}^{t_0}\phi(u)\,du
    \ge
    \gamma_0\phi(t_0)
    =
    \frac{\epsilon}{2\sqrt{2\pi}(1-2\eta)}
\end{align*}
Moreover, $\log(1/(1-\rho))\ge\rho$ and $3\sin(\bar\theta/2)/(16\gamma_0)\le1/\gamma_0\le2/\epsilon$, which gives $K+1\le1+\log(2/\epsilon)/\rho=O(\log(1/\epsilon))$. Eq.~\eqref{eq:batch-size} therefore holds with batches of size $m=O((d+\log(1/\delta)+\log\log(1/\epsilon))/((1-2\eta)\epsilon))$, for a total of
\begin{align*}
    N_2
    =
    (K+1)m
    =
    O\left(
        \frac{\log(1/\epsilon)\left(d+\log(1/\delta)+\log\log(1/\epsilon)\right)}{(1-2\eta)\epsilon}
    \right)
    =
    \widetilde O_\eta\left(\frac{d+\log(1/\delta)}{\epsilon}\right)
\end{align*}
samples. These bounds on $K$ and on $P(t_0,\gamma_0)$ do not depend on the inputs, so $N_2$ can be fixed before $w_0$ and $t_0$ are computed. Finally, iteration $k$ computes one inner product and at most one vector addition for each sample of the batch $S^{(k)}$, and every sample belongs to one batch. The runtime is therefore $O(dN_2)$.
\end{proof}
 
Combining the warm start, the threshold guess, and the refinement with a final threshold ERM proves Theorem~\ref{thm:curriculum-rcn-target}, which we restate below.
 
\begin{theorem}[Restatement of Theorem~\ref{thm:curriculum-rcn-target}]
\label{app:thm:curriculum-rcn-target}
Assume the two-stage model of Definition~\ref{def:two-stage-curriculum}. The source stage satisfies
the assumptions of Theorem~\ref{thm:rare-tail-direction-rv} with $p_-^{(1)}=\PP_{D_1}(Y=-1)$, and the target stage is $\eta$-RCN
with known constant $\eta<1/2$. Suppose $t\ge1$ and $p=\Phi(-t)\ge C_\eta\epsilon$. If
\begin{align*}
    n_1
    =
    \widetilde O_{c,C}\left(
        (d+\log(1/\delta))
        \max\left\{
            \frac1p,\,
            \frac{t^4(\log(e+t))^{2C}}{p_-^{(1)}}
        \right\}
    \right)
\end{align*}
source samples and
\begin{align*}
    n_2
    =
    \widetilde O_\eta\left(
        \frac{d+\log(1/\delta)}{\epsilon}
        +
        \frac{1}{\epsilon^2}
    \right)
\end{align*}
target RCN samples are used, then Algorithm~\ref{alg:rare-tail-curriculum-rcn} returns a halfspace
$\widehat h$ satisfying
\begin{align*}
    \operatorname{err}_{D_2}(\widehat h)-\operatorname{err}_{D_2}(f^\star)
    \le
    \epsilon
\end{align*}
with probability at least $1-\delta$. The runtime is $\widetilde O_\eta(d(n_1+n_2))$.
\end{theorem}
 
\begin{proof}
Split the target samples as in Algorithm~\ref{alg:rare-tail-curriculum-rcn}. We first show that the excess risk is at most $C_\star\epsilon$ for a universal constant $C_\star$, and rescale $\epsilon$ at the end.
 
\emph{Warm start.} Apply Theorem~\ref{thm:rare-tail-direction-rv} to the source sample with confidence parameter
$\delta/4$. Increasing the hidden constant in $n_1$ if necessary gives
\begin{align*}
    \angle(\widehat w_0,w^\star)\le \frac{1}{7t}
\end{align*}
with probability at least $1-\delta/4$.
 
\emph{Threshold guess.} Apply Lemma~\ref{lem:rcn-band-threshold-estimation} to $S_{2,\rm band}$ with $a=1/2$ and confidence
parameter $\delta/4$. With
\begin{align*}
    |S_{2,\rm band}|
    =
    O_\eta\left(\frac{1}{\epsilon^2}\log(1/\delta)\right)
\end{align*}
we get
\begin{align*}
    |\widehat t_0-t|\le \frac{\gamma_\star}{2}
\end{align*}
with probability at least $1-\delta/4$.
 
\emph{Refinement.} On the intersection of these two events, the pair $(\widehat w_0,\widehat t_0)$ satisfies the hypotheses of
Theorem~\ref{app:thm:band-scale-local-refinement}, and it is independent of $S_{2,\rm ref}$. Applying that theorem to $S_{2,\rm ref}$ with
confidence parameter $\delta/4$ and
\begin{align*}
    |S_{2,\rm ref}|
    =
    \widetilde O_\eta\left(\frac{d+\log(1/\delta)}{\epsilon}\right)
\end{align*}
gives
\begin{align*}
    \angle(\widehat w_{\rm ref},w^\star)
    \le
    12\gamma_\star
\end{align*}
with probability at least $1-\delta/4$.
 
\emph{Final threshold.} Algorithm~\ref{alg:threshold-erm} is run on the independent split $S_{2,\rm thr}$ along
$\widehat w_{\rm ref}$, and returns an empirical risk minimizer over the thresholds along this direction. We apply Lemma~\ref{lem:threshold-erm-sample} conditionally on $\widehat w_{\rm ref}$, with accuracy $\epsilon$ and confidence parameter $\delta/4$. With
\begin{align*}
    |S_{2,\rm thr}|
    =
    O\left(\frac{1}{\epsilon^2}\left(\log(1/\epsilon)+\log(1/\delta)\right)\right)
\end{align*}
samples, it gives, with probability at least $1-\delta/4$,
\begin{align*}
    \operatorname{err}_{D_2}(\widehat h)
    \le
    \inf_{b\in\mathbb R}
    \operatorname{err}_{D_2}\left(\operatorname{sign}(\langle \widehat w_{\rm ref},\cdot\rangle+b)\right)
    +\epsilon
    \le
    \operatorname{err}_{D_2}(h_t)+\epsilon
\end{align*}
where $h_t(x)=\operatorname{sign}(\langle \widehat w_{\rm ref},x\rangle+t)$ is the halfspace with direction $\widehat w_{\rm ref}$ and the true threshold.
 
\emph{Excess risk.} It remains to bound the excess risk of $h_t$. Under $\eta$-RCN, every classifier $h$ satisfies
\begin{align*}
    \operatorname{err}_{D_2}(h)-\operatorname{err}_{D_2}(f^\star)
    =
    (1-2\eta)\PP(h(X)\ne f^\star(X))
\end{align*}
By eq.~\eqref{eq:band-scale}, after increasing $C_\eta$ if necessary, $12\gamma_\star\le c_0/t$ for the constant $c_0$ of Lemma~\ref{lem:angle-to-risk-fixed-threshold}. This lemma then
gives
\begin{align*}
    \PP\left(
        h_t(X)
        \ne
        f^\star(X)
    \right)
    \le
    C_0\,
    \angle(\widehat w_{\rm ref},w^\star)\,\phi(t)
    \le
    12C_0\gamma_\star\phi(t)
    =
    \frac{6C_0\,\epsilon}{\sqrt{2\pi}(1-2\eta)}
\end{align*}
since $\gamma_\star\phi(t)=\epsilon/(2(1-2\eta)\sqrt{2\pi})$. Hence $\operatorname{err}_{D_2}(h_t)-\operatorname{err}_{D_2}(f^\star)\le6C_0\epsilon/\sqrt{2\pi}$, and
\begin{align*}
    \operatorname{err}_{D_2}(\widehat h)-\operatorname{err}_{D_2}(f^\star)
    \le
    \left(\frac{6C_0}{\sqrt{2\pi}}+1\right)\epsilon
    =:
    C_\star\epsilon
\end{align*}
A union bound over the warm start, the threshold guess, the refinement,
and the final threshold gives success probability at least $1-\delta$. Running the target stage with accuracy parameter $\epsilon/C_\star$ in place of $\epsilon$ gives excess risk at most $\epsilon$. Since $p\ge C_\eta\epsilon\ge C_\eta\epsilon/C_\star$, the assumptions remain satisfied, and $n_2$ changes only by a constant factor.
 
\emph{Sample size and runtime.} The sample bound follows by summing the three parts of the target sample. Algorithm~\ref{alg:rare-tail-direction} makes one pass over the source samples, at cost $O(dn_1)$. The threshold guess uses only the target labels, at cost $O(n_2)$. The refinement costs $O(d\,|S_{2,\rm ref}|)$ by Theorem~\ref{app:thm:band-scale-local-refinement}, and the final threshold ERM costs $O(d\,|S_{2,\rm thr}|+|S_{2,\rm thr}|\log|S_{2,\rm thr}|)$. Taken together, the runtime is $\widetilde O_\eta(d(n_1+n_2))$.
\end{proof}
 
\begin{remark}[Known noise rate]
\label{rem:curriculum-known-eta}
Both target-stage procedures require knowledge of the noise rate. Algorithm~\ref{alg:rcn-band-threshold-from-labels} relies on the calibration identity
\begin{align*}
    \PP(Y=-1)=\eta+(1-2\eta)p
\end{align*}
and Algorithm~\ref{alg:local-direction-refine} uses $\eta$ in its update and in its step sizes. This matches the known-noise-rate setting in which \citet{diakonikolas2023nearoptimalboundslearninggaussian} analyze their optimization procedure. For unknown $\eta$, they run it for every value of $\eta$ on a grid of $\epsilon$-separated values and select among the resulting halfspaces by testing, and they report that this multiplies the runtime by $O(1/\epsilon)$ and does not change the sample complexity beyond constant factors \citep[Appendix~B.3]{diakonikolas2023nearoptimalboundslearninggaussian}. The source stage of our learner does not involve $\eta$, so only the target stage would need such a wrapper.
\end{remark}

\end{document}